%% file: arXiv.tex
\documentclass[runningheads]{llncs}
\usepackage{multirow}
\usepackage{makecell}
\usepackage{comment}
\usepackage{pifont}
\usepackage{enumitem}

\usepackage{xcolor}
\usepackage{xspace}

\newlist{appendixcontents}{enumerate}{1}
\setlist[appendixcontents]{
    label=\textcolor{red}{\textbf{\textsf{\Alph*}}},
    labelsep=1em,
    leftmargin=*,
    itemsep=0.2em,
    format=\bfseries
}
\definecolor{now}{RGB}{218,227,243}
\definecolor{nowlight}{RGB}{240,244,250}
\definecolor{mcqlight}{RGB}{252,228,236}

\definecolor{nowstate}{RGB}{218,227,243}
\definecolor{fgbglight}{RGB}{240,244,250}
\definecolor{bgfglight}{RGB}{226,239,218}

\definecolor{future}{RGB}{218,227,243}
\definecolor{futurelight}{RGB}{240,244,250}
\definecolor{unpredlight}{RGB}{252,228,236}

\definecolor{past}{RGB}{226,240,217}
\definecolor{current}{RGB}{255,242,204}
\definecolor{pastlight}{RGB}{241,247,237}
\definecolor{currentlight}{RGB}{255,249,229}

\definecolor{mygray}{gray}{0.9}

\providecommand{\hc}[2]{}
\renewcommand{\hc}[2]{\cellcolor{#1}\hspace{1.5pt}#2\hspace{1.5pt}}

\usepackage{eccv}

\usepackage{eccvabbrv}

\usepackage{graphicx}
\usepackage{booktabs}
\usepackage{multirow}
\usepackage{makecell}
\usepackage{xcolor,colortbl}
\usepackage{subcaption}

\usepackage[accsupp]{axessibility}  

\usepackage{hyperref}

\usepackage{orcidlink}

\newcommand{\method}{EgoSAT\xspace}

\begin{document}

\title{EgoSAT: A Comprehensive Benchmark of \underline{Ego}centric \underline{S}treaming Inter\underline{a}c\underline{t}ion Understanding} 

\titlerunning{EgoSAT: Egocentric Streaming Interaction Benchmark}
\author{Yijia Lei\inst{1} \and
Jinzhao Li\inst{1}$^\dagger$  \and
Yichi Zhang\inst{1} \and
Jiacheng Hua\inst{1} \and
Yin Li\inst{2} \and
Miao Liu\inst{1}$^\ddagger$
}
\authorrunning{Y.~Lei et al.}

\institute{College of AI, Tsinghua University \and
University of Wisconsin–Madison
\email{\{leiyj23,lijinzha22\}@mails.tsinghua.edu.cn, miaoliu@mail.tsinghua.edu.cn}
}

\maketitle

\begingroup
\renewcommand{\thefootnote}{}
\footnotetext{$^\dagger$ Project Lead. $^\ddagger$ Corresponding author.}
\endgroup

\begin{center}
    \includegraphics[width=0.95\linewidth]{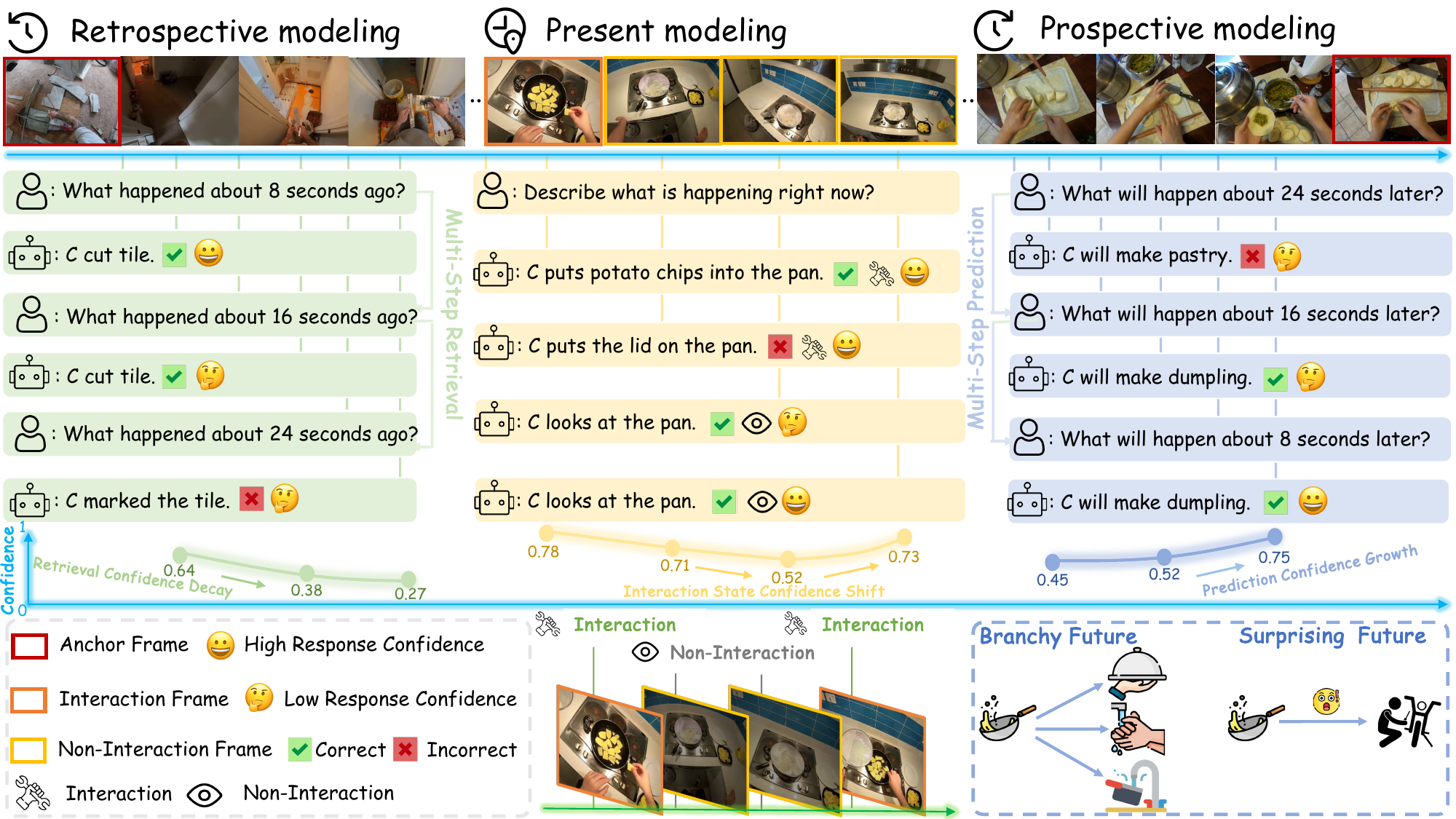}
    \vspace{-0.5em}
    \captionof{figure}{\textbf{\method} presents a unified formulation that brings together several conventional vision–language tasks, \eg, video question answering, online video narration, and activity anticipation, within a single streaming setting. In doing so, it provides the first comprehensive benchmark for evaluating the ability of modern vision–language models to reason about the \textit{past, present, and future} under streaming observations.}\vspace{-0.5em}
    \label{fig:teaser}
\end{center}

\begin{abstract}
  We introduce \textbf{EgoSAT}, the first comprehensive benchmark for egocentric video reasoning in streaming settings, designed to evaluate the capabilities of modern vision–language models (VLMs). The benchmark targets streaming interaction understanding, where video frames arrive sequentially and models must continuously interpret evolving visual context.
  EgoSAT unifies several previously distinct tasks within a single streaming framework. In this formulation, queries about completed events correspond to retrospective reasoning, queries about ongoing activities require online understanding, and queries about future actions involve prospective anticipation. This unified setting requires models to reason about the past, present, and future while operating under the constraint that only previously observed frames are available.
  EgoSAT contains 1,997 unique videos spanning 165 hours of egocentric footage and around 4,800 high-quality question–answer pairs, carefully designed to probe reasoning across varying temporal contexts. Using this benchmark, we evaluate a diverse set of both open-weight and closed-weight VLMs, providing a systematic assessment of their ability for streaming interaction understanding. By distinguishing answerability and conducting diagnostics on confidence of models, we find existing models not only struggle with prospective and retrospective modeling, but also exhibit severe mis-calibration: confidence often fails to track inherent answerability, leading to dangerous ``confidently wrong'' behaviors. Project page: \textcolor{magenta}{\url{https://leiyj23.github.io/EgoSAT/}}
\end{abstract}


\input{chapters/intro}

\input{chapters/related_work}

\input{chapters/benchmark}

\input{chapters/experiment}

\input{chapters/conclusion}

\section*{Acknowledgments}

This work is supported by Tsinghua University--Keystone Electrical (Zhejiang) Co., Ltd. Joint Research Center for Embodied Multimodal Artificial Intelligence (JCEMAI).

%
%
\bibliographystyle{splncs04}
\bibliography{main}

\clearpage
\appendix
\setcounter{section}{0}
\setcounter{figure}{0}
\setcounter{table}{0}
\renewcommand{\thesection}{\Alph{section}}
\renewcommand{\thefigure}{A\arabic{figure}}
\renewcommand{\thetable}{A\arabic{table}}

\section*{Supplementary Material}
\input{supplement_materials/supplementary_body}

\end{document}

%% file: chapters/intro.tex
\section{Introduction}
\label{sec:intro}

Advances in wearable cameras and edge computing, together with recent breakthroughs in vision language models (VLMs), have ushered in a new generation of AI systems that provide context-aware assistance to camera wearers through natural language interfaces, often combined with voice commands and gesture-based controls. A key characteristic of this setting is \textit{streaming processing}: the AI assistant must continuously perceive, comprehend, and retain both the video captured by the device and queries issued by the user, in order to reason about past events (\eg, video question answering), describe ongoing activities (\eg, online  video narration), and predict future intent (\eg, activity anticipation).

Traditionally, these problems have been studied in silos, with dedicated methods and benchmarks designed for individual tasks. However, evaluating VLMs on isolated tasks provides only a partial assessment of their capabilities in realistic streaming scenarios, where perception and reasoning about past, present, and future events is inherently intertwined. In practice, an AI assistant must seamlessly integrate these capabilities within a unified, temporally evolving context.

\textbf{Our key insight} is that \textit{many previously distinct tasks can be naturally reformulated within a single streaming framework}. 
In this setting, video frames arrive sequentially as a continuous stream, and user queries in text form may occur at arbitrary time points, and models are restricted to using only the portion of the video stream observed up to the time of the query. 
As shown in Figure \ref{fig:teaser}, a query about a completed event (\ie, \textbf{retrospective}) corresponds to standard video question answering (QA); a query about an ongoing event (\ie, \textbf{present}) instantiates online video QA; and a query about a future event (\ie, \textbf{prospective}) falls under event anticipation.

Beyond producing accurate answers, they must infer the temporal context of each query, and determine whether sufficient evidence has been observed. Moreover, they must continuously update their confidence as new evidence accumulates, and appropriately calibrate their uncertainty in light of multiple plausible future outcomes. This unified formulation enables a more realistic and holistic evaluation of streaming video understanding in the context of AI assistants


To this end, we introduce \textbf{EgoSAT}, the first benchmark for \underline{Ego}centric \underline{S}treaming inter\underline{a}c\underline{t}ion understanding, designed to evaluate the temporal reasoning capabilities of modern VLMs. We leveraged Ego4D, a scenario-rich and interaction-dense egocentric video corpus with careully curated temporal annotations. 

Leveraging EgoSAT, we conduct extensive evaluation of both closed- and open- weight MLLMs under a strict online protocol.
Across all settings, we find that prospective anticipation and retrospective retrieval remain challenging even for frontier models, while online streaming baselines further lag behind offline counterparts due to irreversible input-side compression. Finally, our answerability-aware confidence diagnostics reveal that model confidence often fails to track factual answerability, frequently staying high on inherently uncertain or incorrect predictions.

%


\medskip
\noindent \textbf{Our main contributions} are summarized as follows.
\begin{enumerate}[nosep]
    \item We present \textbf{\method, a comprehensive benchmark} for egocentric streaming interaction understanding, designed to evaluate the temporal reasoning capabilities of modern VLMs.
    \item Our key \textbf{technical innovations} are (1) \textit{a unified streaming reasoning formulation} integrates several conventional vision-language tasks and requires models to reason about the past, present and future with streaming observations; and (2) \textit{a principled quantification of answerability and its corresponding benchmark design} for future event prediction, enabling fair and systematic evaluation of VLMs.
    \item Leveraging \method, we conduct \textbf{a systematic evaluation} of a diverse set of both open-weight and closed-weight vision–language models. Our empirical results reveals that (1) \textit{prospective anticipation and retrospective retrieval remain challenging across model families}, and online streaming models further lag behind offline counterparts; (2) \textit{confidence is often poorly calibrated to factual answerability}, with several models remaining confidently wrong, motivating answerability-aware diagnostics beyond accuracy.
\end{enumerate}

%% file: chapters/related_work.tex
\section{Related Work}

\noindent\textbf{Streaming Visual Language Models}.\
Streaming visual language models operate under an online prefix constraint where each query can only use frames observed so far. This setting requires low latency responses, continual state updates, and sustained reasoning over long streams. Vinci presents a real time embodied assistant based on egocentric vision language models~\cite{huang2024vinci}, and Dispider introduces a disentangled perception, decision, and reaction framework for active real time interaction~\cite{qian2025dispider}. Streaming long video understanding with large language models(LLMs) proposes a streaming framework that processes videos segment by segment with memory propagation~\cite{qian2024streaming}, while STREAMCHAT studies streaming video understanding with multiround interaction and memory enhanced knowledge~\cite{xiongstreaming}. Streaming VLM targets real time understanding for infinite video streams and aligns training with streaming inference through compact memory management~\cite{xu2025streamingvlm}. VideoLLM-online introduces the LIVE framework with efficient visual token encoding and decoding for online interaction~\cite{chen2024videollm}, and TimeChat-Online proposes to prune redundant visual tokens in streaming videos~\cite{yao2025timechat}. LiveCC builds a streaming speech transcription pipeline for temporally aligned training and provides a benchmark for streaming video language models~\cite{chen2025livecc}.

These works focus on system design, memory management, and efficiency for continuous streams. In egocentric settings, critical evidence is often localized around hands, objects, and gaze, and evidence can decay when background tokens dominate the budget. Our approach targets this interaction blind gap by prioritizing Region-of-Interest (ROI)-aware token budgeting and by training confidence to track evidence accumulation and decay for real time decisions.

\smallskip
\noindent\textbf{Efficient Token Compression for Long Video Understanding}.\
A large body of work improves efficiency by compressing or selecting visual tokens for long video and multimodal understanding. LongVU explores spatiotemporal adaptive compression for long video understanding~\cite{shen2025longvu}, and PVC proposes progressive visual token compression across frames~\cite{yang2025pvc}. TESTA aggregates temporal and spatial tokens to condense video semantics~\cite{ren2023testa}, and VideoChat-Flash introduces hierarchical compression for long context video modeling~\cite{li2024videochat}. DeCo analyzes visual projector design and decouples compression from semantic abstraction~\cite{yao2024deco}. TokenLearner learns a compact set of informative tokens~\cite{ryoo2021tokenlearner}, and VQToken introduces discrete token representations through vector quantization~\cite{zhang2025vqtoken}. TopV~\cite{yang2025topv}, FastV~\cite{chen2024image}, ToMe~\cite{bolyatoken}, HoliTom~\cite{shaoholitom}, and DART~\cite{wen2025stop} further explore pruning or merging strategies to accelerate inference. These methods primarily optimize general efficiency, while our work emphasizes interaction centered evidence selection in egocentric streaming.

\smallskip
\noindent\textbf{Video Understanding Benchmarks}.\
Several benchmarks have been developed for long video understanding, include LongVideoBench~\cite{wu2024longvideobench}, LVBench~\cite{wang2025lvbench}, and MLVU~\cite{zhou2025mlvu}, which evaluate long context video language reasoning. Online and streaming evaluation includes OVO-Bench~\cite{niu2025ovo}, StreamingBench~\cite{lin2024streamingbench}, IPIBench~\cite{li2026ipibench}, Inf-Streams-Eval from Streaming VLM~\cite{xu2025streamingvlm}, LiveSports-3K from LiveCC~\cite{chen2025livecc}, and Eyes Wide Open which introduces ESTP Bench for proactive egocentric streaming QA~\cite{zhangeyes}. Egocentric datasets provide rich interaction structure, including EPIC-KITCHENS-100~\cite{damen2022rescaling}, Ego4D~\cite{grauman2022ego4d}, Ego4D Goal-Step for hierarchical procedural understanding~\cite{song2023ego4d}, Ego-Exo4D for ego and exo perspective skill understanding~\cite{grauman2024ego} and EgoProx~\cite{li2026egoprox} for egocentric interaction reasoning from a spatial perspective.

Despite this progress, most existing benchmarks emphasize final accuracy under offline access or simplified streaming conditions. Our benchmark builds on these datasets while enforcing streaming constraints in egocentric settings and explicitly evaluating confidence behavior over time. A detailed benchmark comparison is provided in the supplementary appendix.

%% file: chapters/benchmark.tex
\section{Problem Formulation and Benchmark Design}

In what follows, we first present a temporally structured formulation of streaming interaction understanding. We then formalize the conditions under which a response is warranted under partial observability. Finally, we describe our evaluation tasks and their metrics, and present our benchmark construction.


\subsection{Problem Formulation}

Streaming interaction understanding requires MLLMs to produce timely and accurate judgments from continuously arriving video frames. Unlike conventional offline video understanding under full observability, this setting introduces response-timing challenges under partial observability. Specifically, models must determine (1) interaction presence — whether an interaction is occurring at all to warrant responding to the user’s query, (2) the answerability of the situation — whether the inference is feasible given the currently available partial observation, and (3) the confidence level of the response under streaming uncertainty. Importantly, these challenges are inherently tied to distinct temporal formulations of the reasoning tasks.

Formally, we denote a streaming video as a sequence of short video clips $X_{1:t}=\{x_1,\ldots,x_t\}$, where $x_t$ denotes the clip at time step $t$ and $X_{1:t}$ are thus the observed clips up to $t$. Under a standard multiple-choice question (MCQ) answering setting, given a query $Q$ (often in text form) and a candidate set $C$, the MLLM $f_\theta$ produces a response $\hat{A}=\mathcal{F}_\theta(Q, X_{1:t}) \in C$. We consider the three key reasoning tasks in streaming interaction understanding.

\smallskip
\noindent\textbf{Present modeling}.\ The query $Q$ pertains to the current clip $x_t$, and the model is required to address $Q$ using only $x_t$, without accessing to the past clips $X_{1:t-1}$ preceding $x_t$. In this setting, the model must explicitly assess the presence of interaction related to $Q$ in the current clip $x_t$.

\smallskip
\noindent\textbf{Prospective modeling}.\ The query $Q$ concerns a future clip $x_{t+\tau}$ with $\tau>0$, and the model is required to address $Q$ using only the observed clips $X_{1:t}$, without accessing to future clips $X_{t+1:t+\tau}$. Notably, the model must explicitly reason about if reliable prediction is feasible given the current observations.

\smallskip
\noindent\textbf{Retrospective modeling}.\ The query $Q$ refers to a past clip $x_{t-\tau}\in X_{1:t}$, and the model must resolve $Q$ based on the accumulated observations $X_{1:t}$. Naturally, the model is required to identify and retrieve the relevant prior context associated with $x_{t-\tau}$ from the observed history.

\smallskip
Note that we adopt the terms \emph{prospective} and \emph{retrospective} modeling to emphasize the streaming setting, where responses must be generated under partial observability and appropriate response timing must be determined. This distinguishes our formulation from conventional future anticipation or memory retrieval tasks~\cite{grauman2022ego4d}.



\begin{figure}[t]
  \centering
  \includegraphics[width=0.9\columnwidth]{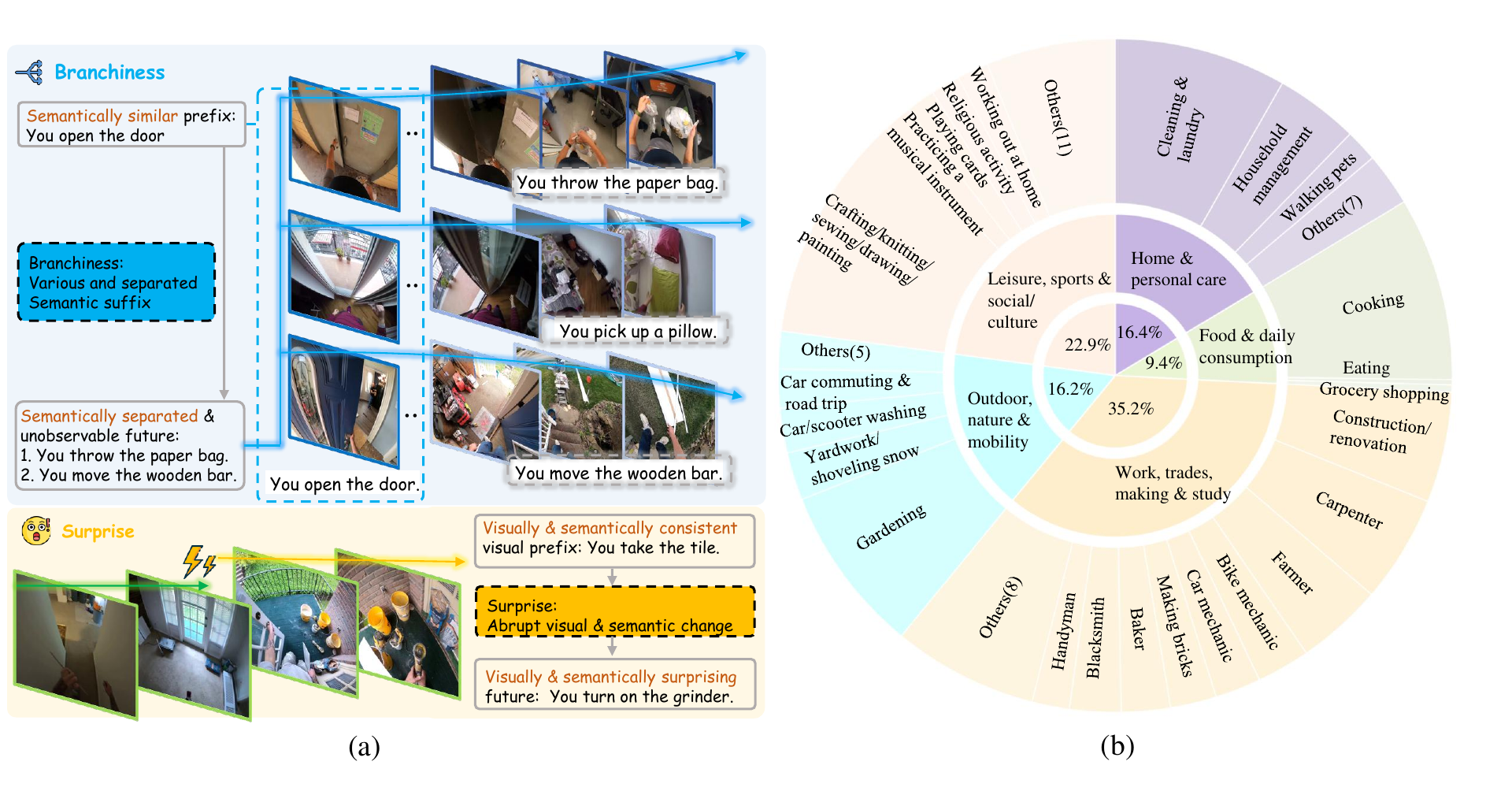}\vspace{-1.5em}
  \caption{Scenario distribution of EgoSAT and examples of predictability.(a) We attribute unpredictability to \textit{branchiness} (a fixed, observable semantic prefix followedd by various and separated semantic suffix) and \textit{surprise} (abrupt visual and semantic change). (b) Our EgoSAT covers 56 different scenarios categorized into five distinct activity groups.}\vspace{-1.5em}
  \label{fig:scenario_stats}
\end{figure}
\subsection{Answerability and Confidence in Prospective Modeling}\label{sec:predictability}

\medskip
\noindent\textbf{Answerability}.\
A key to scientific evaluation in prospective modeling is to distinguish models' capabilities from inherent unpredictability, which derives from partial observation of the activities. Here we assume that the interaction categories and intervals are known, and we carefully characterize the answerability of future events via surprise and branchiness, as shown in Figure~\ref{fig:scenario_stats} (a). We further validate the answerability labels with a human sanity check in the supplementary material, where human judgments agree with our branchiness and surprise labels by 78\% and 84\%, respectively.

\smallskip
\noindent \textbf{(1) Surprise:} \textit{unpredictability induced by abrupt local visual and semantic shifts.}
We quantify \emph{surprising} short-horizon futures by measuring the \textbf{cross-modal discrepancy level} between recent context and the imminent future. Specifically, for each query time $t$, we define a context window $A=[t-\tau,t)$ and a target window with the ground-truth event after $t$: $B=[t,t+h]$. We compute two complementary shift signals.
\textbf{Visual shift:} we extract CLIP image features for frames within $A$ and $B$, average-pool them followed by $\ell_2$-normalize to obtain $v_A$ and $v_B$, and compute the visual similarity $s_v=\cos(v_A,v_B)$. 
\textbf{Semantic shift:} we encode action texts with a CLIP text encoder; for all actions $\{a_i\}$ overlapping $A$, we assign overlap-based weights $w_i$ with $\sum_i w_i=1$, and compare them to the ground-truth text vector $t_B$ via a weighted similarity $s_t=\sum_i w_i\,\cos(t_{a_i},t_B)$. Because $s_v$ and $s_t$ come from different modalities and are not directly comparable in scale, we align them through an empirical distribution transform (probability integral transform + probit). Let $\hat F_v$ and $\hat F_t$ denote the empirical CDFs estimated over all samples; we map
\begin{equation}
p_v=\hat F_v(s_v),\quad p_t=\hat F_t(s_t),\qquad 
z_v=\Phi^{-1}(p_v),\quad z_t=\Phi^{-1}(p_t),
\end{equation}
where $\Phi^{-1}$ is the inverse CDF of the standard normal (probit). We then fuse the two modalities with equal weights,
\begin{equation}
z=(z_v+z_t)/2,\qquad \mathrm{Surprise}(t)=-z,
\end{equation}
where the negative sign ensures that \textbf{lower similarity (stronger mismatch)} corresponds to \textbf{higher surprise} (low percentile $\Rightarrow$ negative $z$). This definition provides a parameter-free, reproducible cross-modal surprise score, enabling answerability-aware stratification and subsequent confidence-based diagnostics.

\smallskip
\noindent \textbf{(2) Branchiness:} \textit{unpredictability induced by diverse and semantically dispersed successor interactions.}
For each query time $t$, let $O$ denote the ongoing interaction covering $t$, and let $A$ be the earliest successor interaction whose start time falls within a short window $(t,t+h]$. Aggregating such transitions over the dataset yields a conditional successor distribution $p(A\mid O)$. We characterize this distribution with two complementary components: (i) \textbf{variety and evenness} via the entropy
\begin{equation}
S(O)=-\Sigma_i\ p_i\log p_i,\quad p_i=p(A_i\mid O),
\end{equation}
and (ii) \textbf{semantic dispersion} via Rao's quadratic entropy
\begin{equation}
Q(O)=\Sigma_i\Sigma_j\ p_i p_j\, d(A_i,A_j),
\end{equation}
where $d(A_i,A_j)=1-\cos(e_i,e_j)$ is the CLIP-text distance between successor interaction texts with their $\ell_2$-normalized embeddings $e_i$. We then define the branchiness score
\begin{equation}
\mathrm{Branchiness}(O)=B(O)=S(O)\cdot Q(O).
\end{equation}
Intuitively, branchiness is high when the future has both many plausible successors (high $S$) and these successors are semantically far apart (high $Q$), indicating low predictability. This measure is complementary to Surprise: while Surprise captures abrupt evidence shifts, Branchiness captures multi-modal, multi-branch continuations, and together they enable answerability-aware evaluation and confidence diagnostics. 

\medskip
\noindent\textbf{Confidence}.\
Unpredictability induced by \textbf{branchiness} and \textbf{surprise}, together with evidence insufficiency in prospective modeling and evidence fade in retrospective modeling, contributes to streaming uncertainty. We therefore introduce \textit{confidence diagnostics} to evaluate whether a model can correctly judge the feasibility of an inference under partial observability.


Following previous work on analyzing multiple-choice confidence scoring for LLMs~\cite{tsvilodub2024predictions}, we quantify MCQ confidence by probing the model’s next-token distribution over the forced one-letter selection set $C$.
Let $p_{\mathrm{raw}}(k)$ be the probability mass assigned to letter $k$ (aggregating its single-token variants), $m=\sum_{k\in C} p_{\mathrm{raw}}(k)$, and $p(k)=p_{\mathrm{raw}}(k)/m$ be the conditional distribution on $C$.
We define confidence and uncertainty as:
\begin{equation}
\mathrm{Conf}=\max\nolimits_{k\in C}\ p(k), \qquad
\mathrm{Ent}=-\Sigma_{k\in C}\ p(k)\log p(k),
\end{equation}
where a low $\mathrm{Conf}$ (or high $\mathrm{Ent}$) indicates low feasibility and enables predictability-/retrievability-aware confidence diagnostics in streaming evaluation.

\begin{figure}[t]
  \centering
  \includegraphics[width=0.9\columnwidth]{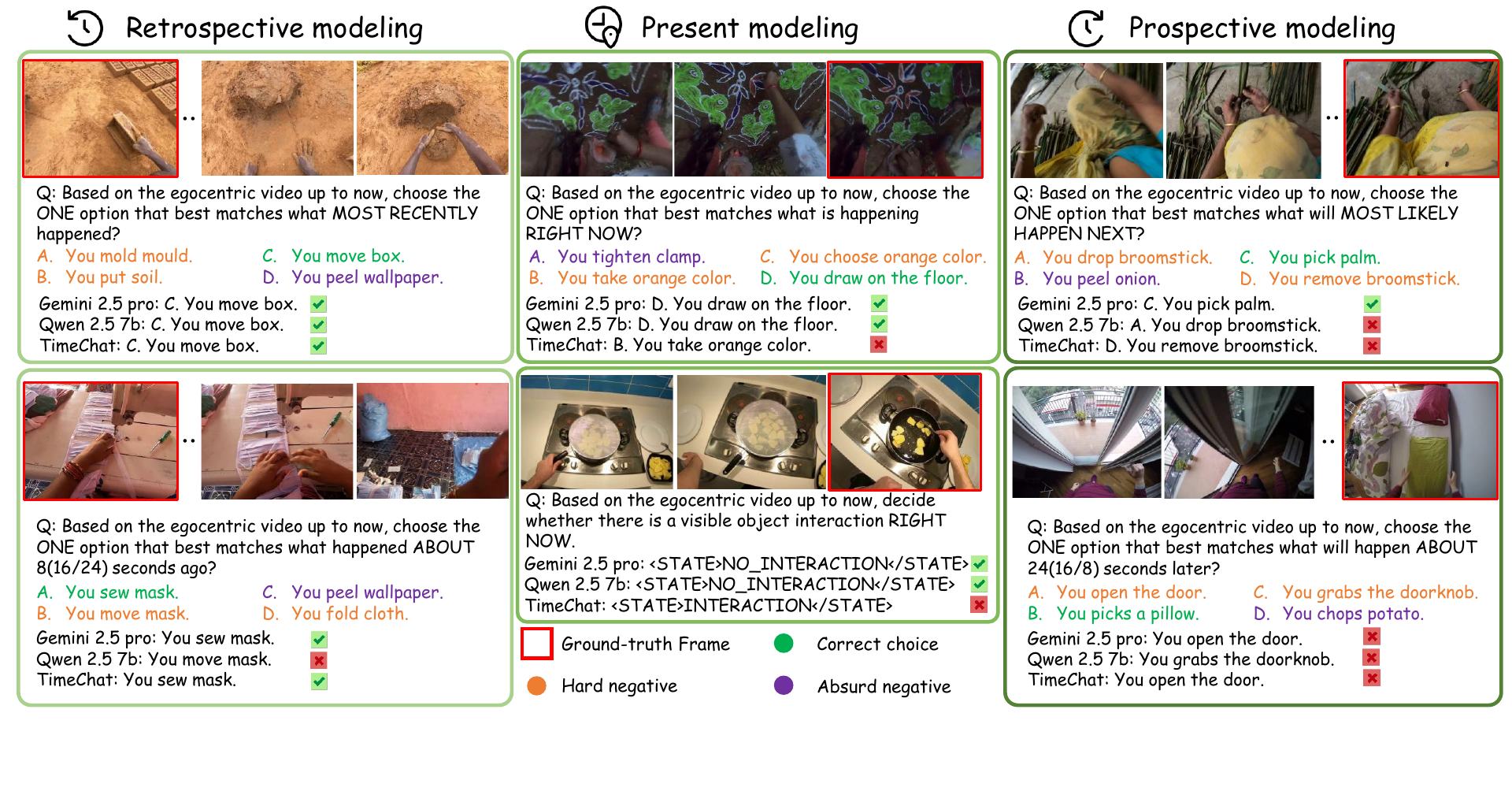}\vspace{-2em}
  \caption{Samples from our six task settings. Red border: the ground-truth frame. Choices are colored accordingly. Notably, the MCQ choices are \textit{shuffled} to ensure minimal information leakage. We also show the model outputs.}\vspace{-1.5em}
  \label{fig:visualization_examples}
\end{figure}
\subsection{Task Formulation and Metric Design}\label{sec:task_metrics}
We now present the evaluation tasks under our problem formulation. For most tasks, we use multiple-choice accuracy as the primary metric to ensure objective and deterministic evaluation. We additionally report open-ended evaluation using LLM-based judging in the supplement. Implementation of these metrics is described in Sec.\ 4.2.


\medskip
\noindent\textbf{Present modeling tasks}.\
For egocentric streaming assistants, real-time understanding via \textit{present modeling} is essential not only for stable online interaction, but also as the perceptual foundation for retro-/prospective reasoning and memory.
We focus on two tasks to evaluate this capability:
\begin{enumerate}[nosep]
    \item \textbf{Now narration}: Recognize the visibility of human-object interaction and what the interaction is.
    \item \textbf{State switch}: Promptly adjust output state when the interaction state switches (interaction to no-interaction, or the reversal).
\end{enumerate}

\medskip
\noindent\textbf{Prospective modeling tasks}.\
Under partial visibility and uncertainty, \textit{prospective modeling} presents significant challenges to streaming assistants. We introduce two tasks that evaluate the capability of utilizing partial observation for future prediction.
\begin{enumerate}[nosep]
    \item \textbf{Short-horizon anticipation}: Given streaming clip at time step $t$, reason over the observed visual prefix to predict the interaction that will occur in the near future at $t+\tau$.
    \item  \textbf{Multi-step anticipation:} Assuming an action begins at $t+\tau$, we probe whether the model can anticipate this event from progressively earlier visual prefixes ending at $t$, $t-\tau$, and $t-2\tau$.
\end{enumerate}


\medskip
\noindent\textbf{Retrospective modeling tasks}.\
Retrospective modeling based on observed history, is another vital demonstration of streaming interaction understanding. We consider the following tasks for evaluating this capability.
\begin{enumerate}[nosep]
    \item \textbf{Short-horizon retrieval}: Retrieve information about the past clip $x_{t-\tau}$ from the streaming inputs
    \item \textbf{Multi-step retrieval}: Retrieve an anchored past event at several past steps forming an arithmetic sequence: $t-\tau$, $t-2\tau$, and $t-3\tau$.
\end{enumerate}


\subsection{Benchmark Construction}
\noindent\textbf{Video source}.\
Our benchmark is built on the recent effort in creating large-scale egocentric video datasets. We adopt Ego4D~\cite{grauman2022ego4d} as the video source. Ego4D is a massive-scale egocentric video dataset with over 3,600 hours of video collected around the world, covering a wide range of (>100) real-world activities and providing human annotated intervals for many hand-object interactions. 

\medskip
\noindent\textbf{Data curation}.\
In an egocentric setting, we define a segment as an interaction segment only when there is a visible hand-object interaction in the view; otherwise, we treat it as a \textit{gap}, \ie, a background segment between interactions. We leverage Ego4D's mature rejection tags as a quality-control mechanism and filter out events that do not contain visible interactions. Specifically, we use two rejection reasons: (1) \textit{no human-object interaction from camera wearer}, and (2) \textit{object of change is not visible}. We validate the reliability of these rejection tags in the appendix through manual inspection.
Finally, we orchestrate 1,997 video sessions with valid annotations, spanning 56 scenarios. Each session lasts roughly 230--300 seconds, with an average duration of 296.8 seconds. These sessions typically contain dense hand-object interactions, making them adequate for constructing streaming QA with interaction-centric ground truth.



\medskip
\noindent\textbf{QA pair construction}.\
Since our QA ground truth is derived from Ego4D annotations and filtered using the above quality-assurance rules, we construct queries using fixed templates with dynamically generated answer options for MCQ evaluation. We additionally provide an open-ended version of all queries using the same templates, enabling complementary evaluation when needed.

Each MCQ instance contains four options: the ground-truth answer, two hard negatives, and one absurd negative. The hard negatives are designed to be temporally confusing and are sampled from events adjacent to the ground-truth segment while ensuring different \textit{(verb, noun)} pairs. Specifically, for Now Narration task, negatives are drawn from the nearest neighboring events of the ground-truth event. For Short-horizon Anticipation and Retrieval tasks, we sample negatives around the \textit{(query time, ground-truth segment)} pair.
For Multi-step Anticipation and Retrospection, negatives are sampled near the query timestamp, as queries and answers may be temporally far apart.

The absurd negative is designed to be semantically distant from the ground truth. We construct a candidate pool of \textit{(verb, noun)} pairs from the training split for each task, remove the top 10\% most frequent pairs to avoid trivial options, and exclude pairs appearing in the same video as the ground truth. The absurd negative is then selected as the candidate with the largest text-embedding distance to the ground-truth option (see supplement for details).

Finally, we randomly shuffle the four options for each MCQ instance and store the shuffled queries offline, following prior benchmark practices to mitigate answer-position bias and potential information leakage. We also include a \textit{blind baseline} using a text-only model to verify that the answer options alone provide little information beyond random guessing.

\medskip
\noindent\textbf{\method}.\ Our final benchmark, EgoSAT, contains 1,997 unique videos spanning 165 hours of egocentric footage and around 4,800 high-quality question–answer pairs. The distribution of videos across scenarios are shown in  Figure~\ref{fig:scenario_stats} (b). In terms of task-level distribution, our benchmark is evenly distributed by design.

%% file: chapters/experiment.tex
\section{Experiments and Results}

\subsection{Experiment Protocol}
\noindent\textbf{Models}.\ We evaluate three groups of models (Table 1).
\begin{enumerate}[nosep]
    \item \textit{Offline video LLMs}, including two proprietary models (Gemini 2.5 Pro~\cite{google2025geminimodelcard}, Claude Sonnet 4~\cite{anthropic2025claude}) and several open-weight models (Qwen2.5-VL~\cite{bai2025qwen25vltechnicalreport}, Video-LLaVA~\cite{lin2024videollavalearningunitedvisual}).
    \item \textit{Streaming VLM}, TimeChat-Online-7B~\cite{yao2025timechat}, which performs streaming inference via dynamic KV cache dropping, Flash-VStream~\cite{flashvstream}, and LLaVA-OV1.5~\cite{llava0v15}.
    \item \textit{ROI-augmented Streaming VLM}, our variant (ROI+TimeChat-Online) of TimeChat-Online that uses egocentric hand and gaze cues for token selection.
\end{enumerate}

\smallskip
\noindent\textbf{Offline-online emulation}.\ Since most off-the-shelf video LLMs only support offline processing, and  do not accept streaming inputs, we simulate strict-online evaluation by truncating the video and retain only the visual prefix at each query time \textit{t}. This ensures no look-ahead for all offline baselines.

\smallskip
\noindent\textbf{Our training-free ROI module}.\ One of the models evaluated in our study, TimeChat-Online, introduces Dynamic Token Drop (DTD) to improve efficiency in streaming settings by mitigating temporal and spatial redundancy in visual tokens. We observe that, in egocentric streaming inference, informative visual regions often concentrate around hand–object interactions and gaze-related areas. Motivated by this observation, we define an Interaction Region of Interest (ROI) that captures the spatial neighborhoods surrounding these egocentric cues.
Building on this idea, we introduce ROI-TimeChat-Online, an ROI-aware variant of TimeChat-Online that prioritizes KV cache of tokens within the Interaction ROI during token selection. Importantly, this modification is parameter-free and training-free, making it easy to incorporate into existing models. We evaluate ROI-TimeChat-Online on our benchmark to study how egocentric priors can improve streaming interaction understanding.



\smallskip
\noindent\textbf{Supervised fine-tuning (SFT)}.\ Beyond zero-shot evaluation, we apply lightweight LoRA-based SFT to the two TimeChat-Online variants using next-token LM loss on EgoSAT target outputs. This setting is used to reduce schema violations and examine whether the observed ROI-related trends persist after task-format alignment.

\smallskip
\noindent\textbf{Metrics}.\ We instantiate the metrics designed in Sec.\ \ref{sec:task_metrics}. Further details can be found in the supplement. \textit{Now Narration Task}: we also report the recall for interaction-visible class. \textit{State Switch Task}: we anchor the query time before switch and gradually moving the After-switch query to the GT Switch moment. Here, a successful state switch occurs if and only if the model predicts correct states in both the two queries. We then report the rate of success when the After-switch query is furthest to the GT switch moment. \textit{Short-horizon Anticipation Task}: we compare confidence scores from the models against those derived in Sec. \ref{sec:predictability}.  \textit{Multi-step Anticipation Task}: we divide queries into three groups based on their relative distance to anchor events: lead $\tau$($2\tau$, $3\tau$) means GT event is $\tau$($2\tau$, $3\tau$) seconds ahead ($\tau=8$), and break down the MCQ accuracy. For \textit{Short-horizon / Multi-step Retrieval Tasks}, we report metrics similar to Prospective Modeling, except that we do not associate answerability with Short-horizon Retrospection in our task settings.

\input{table/table_EXP1}




\subsection{Results on EgoSAT Benchmark}

Table~\ref{tab:main_results} reports the main results across present, prospective, and retrospective modeling tasks. Overall, proprietary frontier models such as Gemini~2.5~Pro and Claude~Sonnet~4 achieve the strongest performance on present narration tasks, suggesting that large-scale pretraining and proprietary data pipelines still provide general advantages even in the challenging egocentric scenarios considered in our benchmark. Interestingly, proprietary models do not exhibit clear advantages on the precision metric for narration generation. We attribute this behavior to human preference alignment objectives, which encourage models to produce reasonable responses even in partially unanswerable settings, leading to lower precision under the strict evaluation protocol adopted in our benchmark.

In addition, both prospective and retrospective modeling pose significant challenges for prevailing methods, albeit in different ways. Retrospective reasoning requires models to retrieve relevant information at the correct temporal point, which can be hindered by imperfect temporal grounding, information loss during sampling, and positional bias introduced by temporal embeddings. On the other hand, prospective modeling introduces inherent uncertainty, as the future outcome often represents only one of many plausible possibilities. This observation further emphasizes the importance of explicitly accounting for predictability in anticipation settings.

\medskip
\noindent\textbf{Online vs.\ offline modeling}.\ We further compare streaming online TimeChat-Online-7B models with offline models of similar scale. Despite operating at comparable model capacity, the online model consistently lags behind its offline counterpart across most tasks. This performance gap primarily stems from the streaming memory constraint: online models must maintain a compressed representation of past observations through incremental caching, which inevitably leads to information loss. As expected, the gap becomes more pronounced in multi-step settings, where longer temporal dependencies exacerbate the information loss accumulated during caching.

\medskip
\noindent\textbf{ROI-based KV caching and task-specific SFT}.\  For a fair comparison, we enforce the same KV caching budget for both TimeChat-Online and its ROI-augmented version. Interestingly, this simple ROI caching improves model performance on present modeling tasks by 3.14\%. However, it leads to large performance degradation on both prospective and retrospective modeling tasks. We attribute this to the fact that these tasks often require reasoning over broader contextual cues in the background to address temporal dynamics, whereas the hand- and gaze-conditioned ROI regions may omit such cues. 

Unsurprisingly, task-specific SFT yields notable performance improvements across tasks, especially for prospective modeling. We freeze the visual encoder during SFT following standard practice. The substantial gains suggest that existing models are already capable of extracting temporally sensitive visual representations, but lack the instruction-level alignment needed to connect these representations with temporally grounded MCQ reasoning tasks. Lightweight SFT can effectively address this misalignment.\

For SFT variants, some entries are \textit{omitted} in Table~\ref{tab:main_results} because the MCQ-first SFT training set only has MCQ type data with structured output schema, which can catastrophically violate the state-only output format required by the interaction-visibility evaluation in Now State Switch. We therefore report additional results for this metric in the supplementary.

\subsection{Answerability Diagnostics}
\input{table/table_multistep_conf}
\noindent\textbf{Confidence-level for state switch task}.\
Interaction boundaries pose a key challenge in streaming settings, where models must switch their output state according to interaction visibility. As shown in Table~\ref{tab:main_results}, although proprietary models achieve relatively better state-switch accuracy, all tested models remain far from satisfactory. We provide a fine-grained analysis of query timing around the ground-truth switch moment in the supplementary material.

\medskip
\noindent\textbf{Confidence-level for multi-step tasks}.
For multi-step anticipation and retrieval, answerability is closely tied to temporal distance. Larger anticipation lead implies less relevant observation for predicting the anchored event, while larger retrieval lag makes past evidence more likely to be diluted or overwritten by later content. We therefore expect both accuracy and self-contained confidence to decrease as lead or lag grows. To test this, we report accuracy at three temporal distances and summarize the confidence--distance relationship by fitting a line to each model's three-point confidence curve, with the normalized mean slope reported in Table~\ref{tab:conf_slope_ms}.
\input{table/table_sh_pred_conf}
Across tasks, multi-step retrospection is generally easier than multi-step anticipation, since the queried events have already occurred and only need to be localized in the observed history. However, many models still fail to show monotonic accuracy degradation as lead/lag increases, especially for anticipation, where local visual similarity and inherent uncertainty can dominate temporal distance. Moreover, confidence slopes are often inconsistent with the expected negative trend, indicating that current models' self-contained confidence does not reliably track factual answerability or temporal difficulty.

\medskip
\noindent\textbf{Predictability}.\ As discussed earlier, predictability for short-horizon anticipation is shaped by the \textit{branchiness} and \textit{surprise} of interaction dynamics. For an uncertainty-aware AI agent, it should assign lower confidence to unpredictable queries than to predictable ones. Note that we evaluate predictability only on short-horizon tasks, as the multi-step setting is already sufficiently challenging; therefore, we restrict those tasks to predictable scenarios. We present the breakdown of predictable and unpredictable queries in Table~\ref{tab:sh_pred_conf}. Not surprisingly, models generally achieve higher accuracy on predictable queries than on unpredictable ones. However, we observe little difference in the confidence levels assigned to these two categories. These results suggest that current MLLMs struggle to distinguish predictable from inherently uncertain situations, often producing overconfident responses even when predictions are incorrect, indicating that their self-contained confidence does not reliably reflect the factual predictability of the query under high unanswerability.




%% file: table/table_EXP1.tex
\definecolor{past}{RGB}{226,240,217}     
\definecolor{current}{RGB}{255,242,204}  
\definecolor{future}{RGB}{218,227,243}   

\definecolor{pastlight}{RGB}{241,247,237}
\definecolor{currentlight}{RGB}{255,249,229}
\definecolor{futurelight}{RGB}{240,244,250}


\begin{table*}[t]
\centering
\scriptsize
\setlength{\tabcolsep}{2pt}  
\renewcommand{\arraystretch}{1.15}
\caption{\textbf{Main results (MCQ-first).} All models are evaluated with strict-online prefix truncation at each query time $t$. \textbf{Rec.} is recall for interaction-visible ($\mathrm{TP}/(\mathrm{TP}+\mathrm{FN})$), and \textbf{Prec.} is precision ($\mathrm{TP}/(\mathrm{TP}+\mathrm{FP})$) \textbf{Multi.} reports average MCQ accuracy over three probes at $\tau,2\tau,3\tau$. }
\label{tab:main_results}

\begin{tabular}{lccccccccc}
\toprule

\noalign{\global\setlength{\tabcolsep}{0pt}}
\multirow{3}{*}{\makecell[l]{Model$\backslash$task}} &
\multicolumn{5}{c}{\hc{current}{Present }} &
\multicolumn{2}{c}{\hc{future}{Prospective }} &
\multicolumn{2}{c}{\hc{past}{Retrospective }} \\

& \multicolumn{3}{c}{\hc{current}{Narration}}
& \multicolumn{2}{c}{\hc{current}{State switch}}
& \multicolumn{1}{c}{\hc{future}{Short}}
& \multicolumn{1}{c}{\hc{future}{Multi.}}
& \multicolumn{1}{c}{\hc{past}{Short}}
& \multicolumn{1}{c}{\hc{past}{Multi.}} \\
\cmidrule(lr){2-10}

& \hc{currentlight}{\makecell[c]{Rec.}}
& \hc{currentlight}{\makecell[c]{Prec.}}
& \hc{currentlight}{\makecell[c]{MCQ\\acc}}
& \hc{currentlight}{\makecell[c]{Fg$\rightarrow$bg}}
& \hc{currentlight}{\makecell[c]{Bg$\rightarrow$fg}}
& \hc{futurelight}{\makecell[c]{MCQ\\acc.}}
& \hc{futurelight}{\makecell[c]{MCQ\\avg. acc.}}
& \hc{pastlight}{\makecell[c]{MCQ\\acc.}}
& \hc{pastlight}{\makecell[c]{MCQ\\avg. acc.}}\\
\midrule

\noalign{\global\setlength{\tabcolsep}{1.5pt}}

\multicolumn{10}{l}{\textit{Human agents}}\\
Human
& 84.62 & 93.13 & 73.13 & 60.63 & 63.75 & 70.63 & 63.13 & 76.25 & 73.75 \\
\midrule
\multicolumn{4}{l}{\textit{Offline  proprietary models}} \\ 
Gemini 2.5 Pro
& 69.63 & 89.02 & 43.00 & 21.43 & 16.67 & 29.92    & 29.43 & 39.80 &  48.15 \\
Claude Sonnet 4
& 77.51 & 86.30 & 38.43 & 11.59 & 12.00 & 32.20 & 29.98  & 40.46 & 39.64  \\
\midrule
\multicolumn{4}{l}{\textit{Offline open-sourced models}} \\ 
Qwen2.5-VL-72B
& 73.66 & 85.64 & 38.86 & 3.95 & 8.64 & 43.06 & 26.00 & 51.87 & 43.46 \\
Qwen2.5-VL-32B
& 86.16 & 86.94 & 39.91 & 2.63 & 7.41 & 37.24 & 25.23 & 50.86 & 40.74 \\
Qwen2.5-VL-7B
& 78.72 & 84.78 & 36.50 & 4.00 & 6.25 & 27.84 & 28.11 & 42.79 & 34.46 \\
Video-LLaVA 7B
& 100.00 & 83.46 & 25.75 & 0.00 & 0.00 & 25.11 & 25.50 & 20.89 & 25.00 \\

\midrule
\multicolumn{10}{@{}l@{}}{\textit{Blind model: see Table 4: Blind Model}}\\
\midrule
\multicolumn{4}{l}{\textit{Online models}} \\ 
TimeChat-Online-7B
& 98.51 & 83.35 & 33.33 & 0.00 & 0.00 & 29.72 & 26.01 & 37.84 & 31.63 \\

ROI-TimeChat-Online
& 97.92 & 83.40 & 36.47 & 0.00 & 1.25 & 30.51 & 25.05 & 34.41 & 30.82 \\

Flash-VStream & 94.61 & 87.44 & 36.92 & 1.02 & 1.78 & 29.64 & 28.50 & 39.76 & 33.80 \\

LLaVA-OV1.5 & 93.27 & 84.68 & 32.30 & 2.54 & 2.03 & 25.53 & 23.68 & 37.84 & 32.39 \\

\midrule
\multicolumn{10}{@{}l@{}}{\textit{SFT Online Models}}\\

TimeChat-Online-7B
& 0.00 & 0.00 & 45.59 & 0.00 & 0.00 & 61.63 & 42.79 & 43.90 & 45.43\\
ROI-TimeChat-Online
& 0.00 & 0.00 & 46.94 & 0.00 & 0.00 & 59.54 & 40.55 & 44.00 & 41.24 \\


\bottomrule
\end{tabular}
\end{table*}

%% file: table/table_multistep_conf.tex

\begin{table*}[t]
\centering
\scriptsize
\setlength{\tabcolsep}{3pt}  
\renewcommand{\arraystretch}{1.15}
\caption{\textbf{Detailed results of prospective and retrospective tasks}. We report confidence slope and accuracy by temporal distance. \textit{Conf slope}: mean least-squares slope of confidence vs.\ temporal distance over the three probes; a negative slope implies confidence decreases with the temporal distance.}
\label{tab:conf_slope_ms}

\resizebox{1.0\textwidth}{!}{%
\begin{tabular}{lcccccccc}
\toprule

\noalign{\global\setlength{\tabcolsep}{0pt}}
\multirow{2}{*}{\makecell[l]{Model}} &
\multicolumn{4}{c}{\hc{future}{Multi-step anticipation}} &
\multicolumn{4}{c}{\hc{past}{Multi-step retrospection}} \\
\cmidrule(lr){2-9}

& \hc{futurelight}{\makecell[c]{Acc\\lead $3\tau$}}
& \hc{futurelight}{\makecell[c]{Acc\\lead $2\tau$}}
& \hc{futurelight}{\makecell[c]{Acc\\lead $\tau$}}
& \hc{futurelight}{\makecell[c]{Conf slope }}
& \hc{pastlight}{\makecell[c]{Acc\\lag $\tau$}}
& \hc{pastlight}{\makecell[c]{Acc\\lag $2\tau$}}
& \hc{pastlight}{\makecell[c]{Acc\\lag $3\tau$}}
& \hc{pastlight}{\makecell[c]{Conf slope}} \\
\midrule

\noalign{\global\setlength{\tabcolsep}{1.5pt}}

{\fontsize{6}{7}\selectfont Qwen-2.5-VL-72B} & 27.99 & 23.51 & 26.49 &-0.76 & 47.78 & 42.22 & 40.37 & -0.16 \\

{\fontsize{6}{7}\selectfont Qwen-2.5-VL-32B} & 25.68 & 22.07 & 27.93 & <0.01 & 42.96 & 40.37 & 38.89 & <0.01 \\

{\fontsize{6}{7}\selectfont Qwen-2.5-VL-7B}  & 27.86 & 29.39 & 31.30 & 0.16 & 40.00 & 37.78 & 37.41 & 0.86 \\

{\fontsize{6}{7}\selectfont Video-LLaVA-7B} & 24.63 & 23.13 & 26.49 & 1.24 & 28.15 & 24.81 & 24.81 & 1.44 \\
\midrule
{\fontsize{6}{7}\selectfont TimeChat-Online-7B} & 27.99 & 26.49 & 29.85 & 0.61 & 44.07 & 42.96 & 42.59 & 0.99 \\

{\fontsize{6}{7}\selectfont ROI-TimeChat-Online} & 29.09 & 20.19 & 25.88 & 17.36 & 30.05 & 28.30 & 34.10 & 4.61 \\
\midrule
{\fontsize{6}{7}\selectfont TimeChat-Online-7B (SFT)} & 42.91 & 42.91 & 42.54 & <0.01 & 45.93 & 48.52 & 41.85 & 0.01 \\
{\fontsize{6}{7}\selectfont ROI-TimeChat-Online (SFT)} & 38.06 & 42.16 & 41.42 & -0.07 & 42.47 & 44.02 & 44.40 & 0.40 \\

\bottomrule
\end{tabular}
}%
\end{table*}

%% file: table/table_sh_pred_conf.tex
\definecolor{unpredlight}{RGB}{252,228,236} 

\begin{table*}[t]
\centering
\scriptsize
\setlength{\tabcolsep}{1.5pt}  
\renewcommand{\arraystretch}{1.15}
\caption{\textbf{Detailed results of short-horizon anticipation}: accuracy and confidence statistics on predictable vs. unpredictable queries.}
\label{tab:sh_pred_conf}
\resizebox{1.0\textwidth}{!}{%
\begin{tabular}{lcccccccc}
\toprule

\noalign{\global\setlength{\tabcolsep}{0pt}}
\multirow{3}{*}{\makecell[l]{Model}} &
\multicolumn{8}{c}{\hc{future}{Short-horizon anticipation}} \\
\cmidrule(lr){2-9}

& \multicolumn{4}{c}{\hc{futurelight}{predictable}} &
\multicolumn{4}{c}{\hc{unpredlight}{unpredictable}} \\
\cmidrule(lr){2-9}

& \hc{futurelight}{accuracy}
& \hc{futurelight}{conf}
& \hc{futurelight}{\makecell[c]{conf\_correct}}
& \hc{futurelight}{\makecell[c]{conf\_wrong}}
& \hc{unpredlight}{accuracy}
& \hc{unpredlight}{conf}
& \hc{unpredlight}{\makecell[c]{conf\_correct}}
& \hc{unpredlight}{\makecell[c]{conf\_wrong}} \\
\midrule

\noalign{\global\setlength{\tabcolsep}{1.5pt}}

Qwen2.5VL-72B & 37.56 & 68.43 & 72.07 & 66.24 & 28.11 & 73.00 & 74.02 & 72.61 \\

Qwen2.5-VL-32B & 38.05 & 65.99 & 68.50 & 64.45 & 31.07 & 66.68 & 69.87 & 65.24 \\

Qwen2.5-VL-7B  & 34.31 & 53.01 & 54.90 & 52.03 & 24.26 & 51.26 & 47.58 & 52.44 \\

Video-LLaVa-7B   & 18.37 & 52.24 & 53.12 & 52.04 & 19.23 & 53.91 & 53.89 & 53.92 \\
\midrule
TimeChat-Online-7B & 33.50 & 49.61 & 51.46 & 48.68 & 28.99 & 48.81 & 49.76 & 48.42 \\

ROI-TimeChat-Online         & 32.58 & 49.14 & 48.90 & 49.25 & 27.16 & 48.40 & 46.49 & 49.11 \\
\midrule
TimeChat-Online-7B (SFT)         & 57.24 & 52.16 & 58.53 & 43.62 & 47.04 & 48.82 & 52.97 & 45.13 \\

ROI-TimeChat-Online (SFT)    & 47.10 & 44.34 & 49.75 & 39.52 & 35.46 & 42.28 & 47.57 & 39.38 \\

\bottomrule
\end{tabular}
}%
\end{table*}

%% file: chapters/conclusion.tex
\section{Conclusion}
We introduced \textbf{EgoSAT}, the first comprehensive benchmark for egocentric streaming interaction understanding that unifies \textbf{retrospective}, \textbf{present}, and \textbf{prospective} reasoning under a strict online protocol. Beyond accuracy, EgoSAT evaluates \textit{answerability} through confidence calibration and future-event predictability. Our results reveal persistent gaps between offline and streaming models, showing that current VLMs struggle to maintain temporally grounded representations under constrained KV-cache budgets. We hope EgoSAT serves as a principled testbed for studying streaming video understanding and developing proactive egocentric AI assistants.

%% file: supplement_materials/supplementary_body.tex
This is the supplementary material for the paper ``EgoSAT: A Comprehensive Benchmark of \underline{Ego}centric \underline{S}treaming Inter\underline{a}c\underline{t}ion Understanding''. We organize the content as follows:

\begin{appendixcontents}[start=1]
    \item -- Implementation Details
    \item -- Additional Experiments
    \item -- Comparison with Related Benchmarks
    \item -- Limitations and Future Work
\end{appendixcontents}

\input{supplement_materials/implementation_details}
\input{supplement_materials/additional_experiments}
\input{supplement_materials/related_work}

%% file: supplement_materials/implementation_details.tex
\section {Implementation Details}
\subsection{Training Details}
We conduct two supervised fine-tuning runs with different output schemas: (1) \textbf{MCQ SFT} for multiple-choice tasks with structured <ANS>, <VERB>, <NOUN>, <DESC> outputs, and (2) \textbf{State SFT} for present modeling tasks that require a decision on state: <STATE> INTERACTION <STATE> vs. <STATE> NO\_INTERACTION <STATE>.

\noindent\textbf{Base model}.\ All SFT runs start from the same base checkpoint of TimeChat-Online-7B. We fine-tune two variants: TimeChat-Online and ROI-TimeChat-Online. MCQ SFT trains the model to produce a structured answer aligned with the MCQ options. State SFT is meant for the state-only schema.\\
\noindent\textbf{MCQ SFT.} We perform a mixed-task fine-tuning stage over the MCQ tasks. For each MCQ task, we sample 1,200 training instances from the training split and merge them into a single mixed training manifest. We then fine-tune the model on this combined manifest to improve task-specific alignment for temporally grounded multiple-choice reasoning.\\
\noindent\textbf{STATE MCQ SFT.} To additionally support tasks with non-MCQ output schemas, we perform a second-stage fine-tuning starting from the checkpoint obtained after the first MCQ SFT stage. In this stage, we use 1,200 samples from Now Narration State-only mode and 1,200 samples from State Switch, together with 120 samples from each MCQ task. This design allows the model to retain competence on the MCQ tasks while improving alignment to state-style outputs required by non-MCQ tasks.

\subsection{Elaboration on Evaluation Metrics}
This section provides additional details on how the evaluation metrics used in our tables and analyses beyond MCQ accuracy are derived.\\
\noindent\textbf{Recall and precision for present modeling.} We treat interaction visibility at the query time as a binary classification problem, where the positive class is INTERACTION. We therefore build a confusion matrix and compute $\mathrm{Precision}=\frac{TP}{TP+FP}$ and $\mathrm{Recall}=\frac{TP}{TP+FN}$.\\
\noindent\textbf{Conf\_correct and conf\_wrong.} Following Sec. 3.2 in the main paper, we derive a self-contained confidence score $c$ from the model's MCQ scoring for each sample. We then report the average confidence conditioned on correctness and obtain the conf\_correct/conf\_wrong scores.\\
\noindent\textbf{Confidence slope for multi-step tasks}. For multi-step anticipation and retrospection, we query the model with three consecutive probes anchored to the same event but with different query-anchor temporal distances. When \textbf{all three probes are correct}, we sort them by increasing distance $\tau$, 2$\tau$, and 3$\tau$ with corresponding confidence $c_1$, $c_2$ and $c_3$. We fit a least-squares line $c \approx \alpha d + \beta$ on the three points, and report the mean slope $\alpha$ across anchors as the \emph{confidence slope}. A well-calibrated model is expected to have $\alpha<0$, indicating lower confidence for probes farther away from the anchor event.\\

\subsection{Details on ROI Construction}
In this section, we give supplementary to how we obtain the ROI annotations, including the hand bounding boxes and gaze region, from the egocentric videos of Ego4D.
\subsubsection{Hand Annotation}
The raw annotations of Ego4D contain human-calibrated annotations of hands and object-of-change, which are pitifully sparse and only covered several key frames of each interaction segment. To ensure constant, dense input of our ROI-driven TimeChat model, we automatically extracted hand ROIs using the MediaPipe Hand Landmarker of Google on Ego4D full-scale videos. To validate the robustness and accuracy of the annotated boxes, we conducted a simple hyperparameter gridsearch, leveraging the sparse Ego4D FHO hand annotations on 40 selected intervals covering 8 most common scenarios. With an IoU-based loss integrated with miss penalty, we found and fixed the ideal configuration for full-scale annotation.
\subsubsection{Gaze Annotation}
We used the GLC gaze estimator for egocentric gaze annotation. It's official release includes pretrained weights for Ego4D; therefore, unlike hand bounding boxes, we did not conduct an additional hyperparameter search and instead directly adopted the provided Ego4D checkpoint. We ran GLC offline on each annotated interval and obtained gaze caches containing timestamp, (\textit{x,y}), radius and confidence for downstream ROI-aware inference.

\subsection{Details on ROI-TimeChat Online}
In this section, we will give a detailed description of how we crafted the ROI-aware variant of TimeChat-Online by switching its DTD module to a ROI-prioritized module.
\subsubsection{ROI-aware Token Dropping}
Our ROI variant is built on top of the original DTD-based TimeChat model. The model reserves the original vision and language backbone, takes the same visual and text inputs, while additionally receives a lightweight ROI cache with each frame. The ROI cache specifies the interaction and gaze centric regions. Our key modification takes place before the decoder attention. Instead of applying token dropping uniformly to all visual tokens, which is how DTD works, we make the dropping policy ROI-aware. Concretely, ROI information is passed through the generation interface and used only when the visual tokens are first constructed. We then select token on the visual token sequence before decoding. The tokens inside the ROI are always preserved, while tokens outside the ROI are further compressed using the original DTD module. In this way, the ROI variant changes only the token selection policy. It preserves interaction-relevant regions and compresses non-ROI regions.
\subsubsection{Memory-budget Control.} For streaming evaluation, we control the cache budget and compare methods under the same memory constraint. Specifically, the released TimeChat-Online setting keeps a 6K video-token budget with an approximately 85\% token drop rate, and our ROI-TimeChat-Online variant is run under the same budget, while applying 85\% drop mainly to background tokens.

%% file: supplement_materials/additional_experiments.tex
\section{Additional Experiments}

\subsection{State Switch: Timeliness and Confidence}
\input{supplement_materials/table_state_switch}
In this section, we present more analysis on the switch timeliness of models' output mode, by gradually moving query time closer to the ground-truth state switch moment. Concretely, we fix the query time before state switch, and make queries $t_1$s, $t_2$s, and $t_3$s after state switch in three distinct runs, and derive the correctness curve consequently. Here, we adopt $t_1=1s$, $t_2=2s$ and $t_3=4s$. A successful state switch requires the model to recognize both the before-switch state and after-switch state correctly. The accuracy slope is derived in a similar way to the confidence slope, with least-squares line fitting.\\
Ideally, when the query time before the switch is fixed, a larger temporal distance after the switch should make the new state easier to recognize. Therefore, a well-calibrated model is expected to show a positive accuracy slope, as well as high state-switch accuracy. From Table~\ref{tab:now_state_switch}, only proprietary MLLMs consistently exhibit positive slopes, and their switch accuracy is significantly higher than other models. In contrast, most open-source models fail to show this desired trend. For the Qwen variants, the slope is weak or even negative, especially for the fg$\rightarrow$bg transition, suggesting unstable trend. Video-LLaVA-7B and TimeChat-Online-7B collapse to near-zero performance, indicating that they almost fail to track state switch at all.\\

Another notable phenomenon is the asymmetry between the two directions. For several models, bg$\rightarrow$fg yields a more positive slope than fg$\rightarrow$bg. This suggests that for most models, entering interaction is easier to detect than exiting from it. The SFT results show that hybrid training fails to resolve the problems with the accuracy trend, but it does improves switch accuracy. Overall, these results show that state-switch evaluation reveals not only existing models exhibit huge performance gap in switch accuracy, but they also demonstrate weak or none calibration to a consistent and ideal belief trend.


\subsection{Confidently Wrong and SFT Impact}

\input{supplement_materials/table_with_sft_state_and_human}

\noindent\textbf{Confidence Calibration and Confidently-wrong Errors.}We present in Table~\ref{tab:conf_correct_wrong_ms}, the absolute value of models' confidence, which is supplementary to the \textit{conf slope} in the main paper. Ideally, the model should be less confident when it chooses wrong answers. However, we observe serious violations of this desirable behavior, especially in multi-step anticipation. Several models, including Qwen 2.5 VL 32B/72B and TimeChat-Online exhibit higher confidence when wrong. This indicates \textbf{overconfident errors} when self-contained confidence fails to calibrate with factual correctness. In retrospection, while several models recover the \textit{confidently wrong} behavior, some still remain overconfidently wrong, suggesting the mis-calibration persists.\\
\noindent\textbf{Effect of SFT.} Comparing TimeChat-Online with its SFT variant, SFT substantially reduces confidence on wrong answers while largely preserving confidence on correct ones. A similar trend exists in the ROI variant, where SFT also lowers confidently wrong consistently, although the model turns out more conservative as its correct confidece is lowered as well. Overall, SFT primarily improves confidence calibration by suppressing overconfident errors rather than elevating absolute confidence.

\subsection{Human Baseline}
In Sec. 4.2 of main paper, we see huge performance gap in our task settings, especially Prospective Modeling and Retrospective Modeling, even for proprietary MLLMs. We therefore conduct comparative experiments on human agents to validate the feasibility or upper-bound of our tasks.
We uniformly sample 10\% of the multiple-choice questions from each of our 6 task to form a validation set for human agents.\\
As shown in Table~\ref{tab:main_results}, humans achieve strong performance across present, prospective and retrospective tasks, substantially surpassing all evaluated models. This gap indicates that EgoSAT is far from saturated and that low model performance primarily reflects capability limitations rather than inherent unanswerability of the benchmark. Meanwhile, human accuracy remains lower on state switch and multistep prediction, highlighting interaction-boundary sensitivity and observation insufficiency as key challenges in streaming interaction understanding.
\input{supplement_materials/table_multistep_conf_abs}

\input{supplement_materials/table_blind}
\subsection{MCQ Option Fairness and Validity}
\subsubsection{Blind Models}
In this section, we validate the fairness of our multiple-choice options, by not providing the models with visual input and asking them to answer the MCQ using text input only. To balance comprehensiveness and cost, we choose the representative models of each family, namely Gemini 2.5 Pro, Qwen 2.5 vl 32b and TimeChat Online 7b for this section.
As shown in Table~\ref{tab:blind}, disabling vision leads to a clear and consistent performance drop on the MCQ-based prospective and retrospective tasks. This indicates that the shuffled options alone do not provide a reliable signal and that models cannot reliably inter the correct choice from the candidate texts without observing the video. Therefore, the blind baseline serves as a validation that our MCQ construction does not leak strong cues, supporting the fairness of the multiple-choice setting used throughout EgoSAT.
\subsubsection{Hard- and Absurd-negative validation}
Beyond the blind-model baseline, we further analyze which type of incorrect option is selected when a model answers an MCQ incorrectly. As described in the main paper, each EgoSAT MCQ contains one ground-truth answer, two hard negatives and one absurd negative. The hard negatives are temporally confusing options sampled from events adjacent or temporally related to the ground-truth segment, while the absurd negative is selected to be semantically distant from the ground-truth answer and weakly related to the current visual context.\
Formally, for question \(i\), let \(y_i^\ast\) denote the ground-truth answer, \(\mathcal{H}_i\) denote the set of two hard negatives, \(a_i\) denote the absurd negative, and \(\hat{y}_i\) denote the model prediction. We define the set of valid incorrect MCQ predictions as
\[
\mathcal{W}
=
\left\{
i \mid \hat{y}_i \neq y_i^\ast,\ 
\hat{y}_i \in \mathcal{H}_i \cup \{a_i\}
\right\}.
\]
We then compute the conditional distribution of wrong-option types:
\[
P_{\mathrm{hard}\mid\mathrm{wrong}}
=
\frac{1}{|\mathcal{W}|}
\sum_{i\in\mathcal{W}}
\mathbf{1}\!\left
[\hat{y}_i \in \mathcal{H}_i\right]
,
\quad
P_{\mathrm{absurd}\mid\mathrm{wrong}}
=
\frac{1}{|\mathcal{W}|}
\sum_{i\in\mathcal{W}}
\mathbf{1}\!\left
[\hat{y}_i = a_i\right]
.
\]
Since each question has two hard negatives and one absurd negative, a uniformly random choice among the three incorrect options would yield an expected hard-negative error rate of \(2/3\) and an absurd-negative error rate of \(1/3\).\

\input{supplement_materials/table_absurd_when_wrong}

As shown in Table~\ref{tab:mcq_reshuffle}, when models answer incorrectly, they predominantly select the hard-negative options rather than the absurd option. This suggests that models can usually reject options that are clearly unrelated to the current context, and that their failures mainly arise from confusing temporally adjacent or semantically plausible but factually incorrect alternatives. This provides empirical support for our MCQ design: hard negatives enable fine-grained evaluation of temporal and interaction understanding, while the absurd negative helps identify random guessing or failures in basic contextual judgment.\

As an additional check for option-position bias, we re-shuffle the answer options and repeat the same analysis. The resulting wrong-option distributions remain highly consistent with those under the original shuffle, further indicating that our offline option shuffling does not introduce systematic bias.\

\subsubsection{Human Validation of Answerability Labels}
To validate whether our automatic answerability metrics align with human judgments of predictability and uncertainty, we conduct a human sanity check for the surprise and branchiness labels. Since the predictable vs. unpredictable split is derived from these automatic metrics, we further examine whether the resulting labels are consistent with human judgment.

Specifically, we construct two binary validation settings based on the metric-defined bins: predictable vs. branchy-only, and predictable vs. surprise-only. A human annotator is provided with the same type of visual and narration evidence used by our metrics, and is asked to decide whether each sample is predictable or belongs to the corresponding uncertainty type. We then compare the human decisions with the automatic labels. The human judgments agree with our branchiness labels in \(78\%\) of the cases and with our surprise labels in \(884\%\) of the cases. These results suggest that, although surprise and branchienss are automatic proxy measures, they are reasonably consistent with human-judged unpredictability in streaming interaction understanding.

This validation is intended as a sanity check rather than a replacement for large-scale human annotation. It supports the use of our answerablitiy split by showing that the metric-defined predictable and uncertain samples largely agree with human judgments of temporal uncertainty.

\subsection{Open-ended Demonstrations and Evaluation}
\input{supplement_materials/table_open}
As we've mentioned in the main paper, we additionally report open-ended evaluation for Now Narration in the supplementary to complement the MCQ results and diagnose model reliability under a more natural output setting. Unlike MCQ, models are not required to output an optional letter; instead, they are prompted to directly generate structured \textit{VERB}, \textit{NOUN} and the composed \textit{ACTION}. We adopt a strict matching rule: a pair is counted as correct only when both the predicted verb and noun exactly match the ground truth. In practice, many models, especially smaller ones, frequently violate the expected schema and produce free text, leading to a large portion of outputs that cannot be deterministically parsed. To reduce failures due to formatting, we use a \textbf{teacher LLM} (Gemini-3-Pro) as an intermediate parser: the teacher converts non-standard outputs into the same structured fields when possible, without making correctness judgments. If the teacher still fails to extract valid fields, the output is treated as invalid counted as wrong.\\
Under this strict protocol, Gemini Pro achieves only 4.6\% pair accuracy on 672 GT interaction samples, indicating that open-ended streaming narration remains challenging even for proprietary models. Overall, we treat open-ended results as a complementary diagnostic to reveal failure modes and robustness, rather than an ideal replacement of MCQ-based evaluation.

\subsection{Memory Proxy Sensitivity}

To complement the end-task evaluation and the prefix-truncation protocol used for offline models, we conduct a memory-proxy diagnostic to study how model performance changes with different amounts of available past context. This experiment is intended as a controlled diagnostic proxy, evaluating how a model uses a compressed memory of previous observations and captruing the memory-management challenges of native streaming architectures.\

For each query timestamp, we fix the dense current visual input to a clip of length \(\tau_m\), and vary the duration of the long-range memory context as multiples of \(\tau_m\). The memory context is represented by sparsely sampled keyframes at 0.25 FPS together with a one-sentence caption generated by Gemini 3.1 Pro. Thus, each model receives a fixed dense current clip plus a sparse memory consisting of keyframes and a textual summary over the preceding context. We evaluate representative offline models, Gemini 2.5 Pro and Qwen2.5-VL-7B, under the same MCQ protocol, and vary the sparse memory horizon among \(\tau_m\), \(2\tau_m\), \(3\tau_m\), and \(4\tau_m\). In our experiments, we set \(\tau_m=10\) seconds.\
\input{supplement_materials/table_memory_proxy}
As shown in Table~\ref{tab:memory_proxy_results}, increasing the memory horizon does not lead to uniform gains. For Gemini 2.5 Pro, longer memory slightly improves Now Narration, but the gains are not monotonic across anticipation and retrieval tasks. Qwen2.5-VL-7B shows similarly small and non-monotonic changes across memory configurations. These results suggest that simply appending more sparse past context and captions is not sufficient for robust streaming understanding. Effective streaming models likely require more adaptive memory selection, compression and retrieval mechanisms to determine which past evidence is relevant to the current query.

\subsection{Cross-scenario SFT Validation}
To address the concern that SFT may overfit to the specific templates or data distribution of EgoSAT, we conduct a cross-scenario validation. We split the SFT data into two disjoint subsets, \(A\) and \(B\), according to Ego4D scenario labels, and compare cross-scenario training against in-scenario training on the same target validation set.\

Specifically, we train two SFT variants. In the \(A\rightarrow B\) setting, the model is fine-tuned only on scenario subset \(A\) and evaluated on a held-out validation set from scenario subset \(B\). In the \(B\rightarrow B\) setting, the model is fine-tuned on the training split of scenario subset \(B\) and evaluated on the same held-out \(B\) validation set. Therefore, \(A\rightarrow B\) measures whether SFT transfers to target scenarios unseen during fine-tuning, while \(B\rightarrow B\) serves as an in-scenario adaptation reference.\

\input{supplement_materials/table_cross_scenario}
As shown in Table~\ref{tab:sft_transfer}, the \(A\rightarrow B\) and \(B\rightarrow B\) results are highly comparable for both TimeChat-Online-7B and ROI-TimeChat-Online. Fro TimeChat-Online-7B, the two settings differ by around one percentage point on most tasks, with the largest gap appearing in Multi-step Anticipation. For ROI-TimeChat-Online, the differences are similarly small across all tasks. These results suggest that the SFT gains can transfer to held-out target scenarios. This experiment shows that the observed SFT improvements are not merely due to in-scenario overfitting.


%% file: supplement_materials/table_state_switch.tex
\begin{table*}[t]
\centering
\scriptsize
\setlength{\tabcolsep}{3pt}
\renewcommand{\arraystretch}{1.15}
\caption{\textbf{Detailed results of Now State Switch}. We report transition accuracy at different probing positions and the fitted accuracy slope for foreground-to-background and background-to-foreground state switches.}
\label{tab:now_state_switch}

\resizebox{1.0\textwidth}{!}{%
\begin{tabular}{lcccccccc}
\toprule

\noalign{\global\setlength{\tabcolsep}{0pt}}
\multirow{3}{*}{\makecell[l]{Model}} &
\multicolumn{8}{c}{\hc{nowstate}{Now State Switch}} \\
\cmidrule(lr){2-9}

& \multicolumn{4}{c}{\hc{fgbglight}{Fg$\rightarrow$bg}} &
\multicolumn{4}{c}{\hc{bgfglight}{Bg$\rightarrow$fg}} \\
\cmidrule(lr){2-5}\cmidrule(lr){6-9}

& \hc{fgbglight}{Acc $t_1$}
& \hc{fgbglight}{Acc $t_2$}
& \hc{fgbglight}{Acc $t_3$}
& \hc{fgbglight}{Acc slope}
& \hc{bgfglight}{Acc $t_1$}
& \hc{bgfglight}{Acc $t_2$}
& \hc{bgfglight}{Acc $t_3$}
& \hc{bgfglight}{Acc slope} \\
\midrule

\noalign{\global\setlength{\tabcolsep}{1.5pt}}

{\fontsize{6}{7}\selectfont Gemini 2.5 Pro}
& 8.96 & 7.14 & 21.43 & 4.5836 & 12.33 & 10.14 & 16.67 & 1.7064 \\

{\fontsize{6}{7}\selectfont Claude Sonnet 4}
& 1.35 & 2.67 & 11.59 & 3.5629 & 6.41 & 8.75 & 12.00 & 1.8293 \\

{\fontsize{6}{7}\selectfont Qwen2.5-VL-72B}
& 5.26 & 3.95 & 3.95 & -0.3743 & 4.94 & 8.64 & 8.64 & 1.0571 \\

{\fontsize{6}{7}\selectfont Qwen2.5-VL-32B}
& 2.63 & 0.00 & 2.63 & 0.1879 & 2.47 & 0.00 & 7.41 & 1.9407 \\

{\fontsize{6}{7}\selectfont Qwen2.5-VL-7B}
& 4.00 & 9.33 & 4.00 & -0.3807 & 7.50 & 3.75 & 6.25 & -0.1786 \\

{\fontsize{6}{7}\selectfont Video-LLaVA-7B}
& 0.00 & 0.00 & 0.00 & 0.00 & 0.00 & 0.00 & 0.00 & 0.00 \\
\midrule

{\fontsize{6}{7}\selectfont TimeChat-Online-7B}
& 0.00 & 0.00 & 0.00 & 0.00 & 0.00 & 0.00 & 0.00 & 0.00 \\

{\fontsize{6}{7}\selectfont ROI-TimeChat-Online}
& 1.32 & 0.00 & 0.00 & -0.3771 & 1.25 & 1.25 & 1.25 & 0.00 \\
\midrule

{\fontsize{6}{7}\selectfont TimeChat-Online-7B(SFT)}
& 18.42 & 13.16 & 10.53 & -2.4421 & 5.00 & 5.00 & 6.25 & 0.4464 \\

{\fontsize{6}{7}\selectfont ROI-TimeChat-Online(SFT)}
& 0.00 & 0.00 & 0.00 & 0.00 & 0.00 & 0.00 & 1.25 & 0.4464 \\

\bottomrule
\end{tabular}
}%
\end{table*}

%% file: supplement_materials/table_with_sft_state_and_human.tex
\definecolor{past}{RGB}{226,240,217}     
\definecolor{current}{RGB}{255,242,204}  
\definecolor{future}{RGB}{218,227,243}   

\definecolor{pastlight}{RGB}{241,247,237}
\definecolor{currentlight}{RGB}{255,249,229}
\definecolor{futurelight}{RGB}{240,244,250}


\begin{table*}[t]
\centering
\scriptsize
\setlength{\tabcolsep}{2pt}  
\renewcommand{\arraystretch}{1.15}
\caption{\textbf{Main results (MCQ-first).} All models are evaluated with strict-online prefix truncation at each query time $t$. \textbf{Rec.} is recall for interaction-visible ($\mathrm{TP}/(\mathrm{TP}+\mathrm{FN})$), and \textbf{Prec.} is precision:($\mathrm{TP}/(\mathrm{TP}+\mathrm{FP})$) \textbf{Multi.} reports average MCQ accuracy over three probes at $\tau,2\tau,3\tau$. }
\label{tab:main_results}

\begin{tabular}{lccccccccc}
\toprule

\noalign{\global\setlength{\tabcolsep}{0pt}}
\multirow{3}{*}{\makecell[l]{Model$\backslash$task}} &
\multicolumn{5}{c}{\hc{current}{Present }} &
\multicolumn{2}{c}{\hc{future}{Prospective }} &
\multicolumn{2}{c}{\hc{past}{Retrospective }} \\

& \multicolumn{3}{c}{\hc{current}{Narration}}
& \multicolumn{2}{c}{\hc{current}{State switch}}
& \multicolumn{1}{c}{\hc{future}{Short}}
& \multicolumn{1}{c}{\hc{future}{Multi.}}
& \multicolumn{1}{c}{\hc{past}{Short}}
& \multicolumn{1}{c}{\hc{past}{Multi.}} \\
\cmidrule(lr){2-10}

& \hc{currentlight}{\makecell[c]{Rec.}}
& \hc{currentlight}{\makecell[c]{Prec.}}
& \hc{currentlight}{\makecell[c]{MCQ\\acc}}
& \hc{currentlight}{\makecell[c]{Fg$\rightarrow$bg}}
& \hc{currentlight}{\makecell[c]{Bg$\rightarrow$fg}}
& \hc{futurelight}{\makecell[c]{MCQ\\acc.}}
& \hc{futurelight}{\makecell[c]{MCQ\\avg. acc.}}
& \hc{pastlight}{\makecell[c]{MCQ\\acc.}}
& \hc{pastlight}{\makecell[c]{MCQ\\avg. acc.}}\\
\midrule

\noalign{\global\setlength{\tabcolsep}{1.5pt}}

\multicolumn{10}{l}{\textit{Human agents}}\\
Human
& 84.62 & 93.13 & 73.13 & 60.63 & 63.75 & 70.63 & 63.13 & 76.25 & 73.75 \\
\midrule
\multicolumn{4}{l}{\textit{Offline  proprietary models}} \\ 
Gemini 2.5 Pro
& 69.63 & 89.02 & 43.00 & 21.43 & 16.67 & 29.92    & 29.43 & 39.80 &  48.15 \\
Claude Sonnet 4
& 77.51 & 86.30 & 38.43 & 11.59 & 12.00 & 32.20 & 29.98  & 40.46 & 39.64  \\
\midrule
\multicolumn{4}{l}{\textit{Offline open-sourced models}} \\ 
Qwen2.5-VL-72B
& 73.66 & 85.64 & 38.86 & 3.95 & 8.64 & 43.06 & 26.00 & 51.87 & 43.46 \\
Qwen2.5-VL-32B
& 86.16 & 86.94 & 39.91 & 2.63 & 7.41 & 37.24 & 25.23 & 50.86 & 40.74 \\
Qwen2.5-VL-7B
& 78.72 & 84.78 & 36.50 & 4.00 & 6.25 & 27.84 & 28.11 & 42.79 & 34.46 \\
Video-LLaVA 7B
& 100.00 & 83.46 & 25.75 & 0.00 & 0.00 & 25.11 & 25.50 & 20.89 & 25.00 \\

\midrule
\multicolumn{10}{@{}l@{}}{\textit{Blind model: see Table 4: Blind Model}}\\
\midrule
\multicolumn{4}{l}{\textit{Online models}} \\ 
TimeChat-Online-7B
& 98.51 & 83.35 & 33.33 & 0.00 & 0.00 & 29.72 & 26.01 & 37.84 & 31.63 \\

ROI-TimeChat-Online
& 97.92 & 83.40 & 36.47 & 0.00 & 1.25 & 30.51 & 25.05 & 34.41 & 30.82 \\

\midrule
\multicolumn{10}{@{}l@{}}{\textit{SFT Online Models}}\\

TimeChat-Online-7B
& 0.00 & 0.00 & 45.59 & 0.00 & 0.00 & 61.63 & 42.79 & 43.90 & 45.43\\
ROI-TimeChat-Online
& 0.00 & 0.00 & 46.94 & 0.00 & 0.00 & 59.54 & 40.55 & 44.00 & 41.24 \\


\bottomrule
\end{tabular}
\end{table*}

%% file: supplement_materials/table_multistep_conf_abs.tex

\begin{table*}[t]
\centering
\scriptsize
\setlength{\tabcolsep}{3pt}
\renewcommand{\arraystretch}{1.15}
\caption{\textbf{Detailed correct/wrong breakdown for multi-step prospective tasks}. We report the number (or percentage) of correct and wrong predictions at different temporal distances for multi-step anticipation and multi-step retrospection.}
\label{tab:conf_correct_wrong_ms}

\resizebox{1.0\textwidth}{!}{%
\begin{tabular}{lcccccccccccc}
\toprule

\noalign{\global\setlength{\tabcolsep}{0pt}}
\multirow{3}{*}{\makecell[l]{Model}} &
\multicolumn{6}{c}{\hc{future}{Multi-step anticipation}} &
\multicolumn{6}{c}{\hc{past}{Multi-step retrospection}} \\
\cmidrule(lr){2-7}\cmidrule(lr){8-13}

& \multicolumn{2}{c}{\cellcolor{futurelight}\makecell[c]{Conf lead $3\tau$}}
& \multicolumn{2}{c}{\cellcolor{futurelight}\makecell[c]{Conf lead $2\tau$}}
& \multicolumn{2}{c}{\cellcolor{futurelight}\makecell[c]{Conf lead $\tau$}}
& \multicolumn{2}{c}{\cellcolor{pastlight}\makecell[c]{Conf lag $\tau$}}
& \multicolumn{2}{c}{\cellcolor{pastlight}\makecell[c]{Conf lag $2\tau$}}
& \multicolumn{2}{c}{\cellcolor{pastlight}\makecell[c]{Conf lag $3\tau$}} \\
\cmidrule(lr){2-3}\cmidrule(lr){4-5}\cmidrule(lr){6-7}
\cmidrule(lr){8-9}\cmidrule(lr){10-11}\cmidrule(lr){12-13}

& \hc{futurelight}{correct}
& \hc{futurelight}{wrong}
& \hc{futurelight}{correct}
& \hc{futurelight}{wrong}
& \hc{futurelight}{correct}
& \hc{futurelight}{wrong}
& \hc{pastlight}{correct}
& \hc{pastlight}{wrong}
& \hc{pastlight}{correct}
& \hc{pastlight}{wrong}
& \hc{pastlight}{correct}
& \hc{pastlight}{wrong} \\
\midrule

\noalign{\global\setlength{\tabcolsep}{1.5pt}}

{\fontsize{6}{7}\selectfont Qwen-2.5-VL-72B}
& 71.73 & 67.73 & 68.14 & 71.81 & 68.66 & 71.66 & 78.34 & 66.41 & 75.12 & 68.26 & 75.74 & 63.77 \\

{\fontsize{6}{7}\selectfont Qwen-2.5-VL-32B}
& 60.31 & 61.15 & 58.25 & 63.01 & 61.16 & 61.73 & 67.44 & 65.95 & 66.87 & 59.27 & 65.40 & 64.17 \\

{\fontsize{6}{7}\selectfont Qwen-2.5-VL-7B}
& 50.81 & 52.85 & 50.70 & 55.10 & 50.82 & 53.32 & 57.49 & 53.44 & 56.74 & 53.86 & 56.83 & 52.34 \\

{\fontsize{6}{7}\selectfont Video-LLaVA-7B}
& 53.87 & 52.82 & 54.74 & 52.77 & 55.20 & 52.77 & 48.81 & 49.57 & 48.73 & 48.84 & 47.38 & 48.88 \\
\midrule

{\fontsize{6}{7}\selectfont TimeChat-Online-7B}
& 48.50 & 49.73 & 47.54 & 51.85 & 47.86 & 51.40 & 53.39 & 50.08 & 51.48 & 49.46 & 50.66 & 49.27 \\

{\fontsize{6}{7}\selectfont ROI-TimeChat-Online}
& 48.22 & 48.54 & 47.41 & 51.10 & 49.30 & 48.52 & 52.58 & 51.37 & 49.97 & 50.22 & 49.14 & 49.65 \\
\midrule

{\fontsize{6}{7}\selectfont TimeChat-Online-7B (SFT)}
& 48.42 & 42.74 & 47.69 & 43.65 & 49.53 & 43.01 & 52.97 & 45.48 & 51.31 & 46.44 & 50.82 & 45.46 \\

{\fontsize{6}{7}\selectfont ROI-TimeChat-Online (SFT)}
& 44.04 & 40.36 & 44.83 & 41.65 & 44.78 & 40.65 & 47.42 & 43.60 & 47.23 & 43.30 & 47.33 & 43.00 \\

\bottomrule
\end{tabular}
}%
\end{table*}

%% file: supplement_materials/table_blind.tex
\definecolor{past}{RGB}{226,240,217}     
\definecolor{current}{RGB}{255,242,204}  
\definecolor{future}{RGB}{218,227,243}   

\definecolor{pastlight}{RGB}{241,247,237}
\definecolor{currentlight}{RGB}{255,249,229}
\definecolor{futurelight}{RGB}{240,244,250}


\begin{table*}[t]
\centering
\scriptsize
\setlength{\tabcolsep}{3pt}  
\renewcommand{\arraystretch}{1.15}
\caption{\textbf{Blind Models (MCQ-first).} We choose three representative models of their families and present the results of their \textit{blind vs. non-blind} version here.}
\label{tab:blind}

\begin{tabular}{lcccccccc}
\toprule

\noalign{\global\setlength{\tabcolsep}{0pt}}
\multirow{3}{*}{\makecell[l]{Model$\backslash$task}} &
\multicolumn{4}{c}{\hc{current}{Present }} &
\multicolumn{2}{c}{\hc{future}{Prospective }} &
\multicolumn{2}{c}{\hc{past}{Retrospective }} \\

& \multicolumn{2}{c}{\hc{current}{Narration}}
& \multicolumn{2}{c}{\hc{current}{State switch}}
& \multicolumn{1}{c}{\hc{future}{Short}}
& \multicolumn{1}{c}{\hc{future}{Multi.}}
& \multicolumn{1}{c}{\hc{past}{Short}}
& \multicolumn{1}{c}{\hc{past}{Multi.}} \\
\cmidrule(lr){2-9}

& \hc{currentlight}{\makecell[c]{Prec.}}
& \hc{currentlight}{\makecell[c]{MCQ\\acc}}
& \hc{currentlight}{\makecell[c]{Fg$\rightarrow$bg}}
& \hc{currentlight}{\makecell[c]{Bg$\rightarrow$fg}}
& \hc{futurelight}{\makecell[c]{MCQ\\acc.}}
& \hc{futurelight}{\makecell[c]{MCQ\\avg. acc.}}
& \hc{pastlight}{\makecell[c]{MCQ\\acc.}}
& \hc{pastlight}{\makecell[c]{MCQ\\avg. acc.}}\\
\midrule

\noalign{\global\setlength{\tabcolsep}{1.5pt}}

\multicolumn{4}{l}{\textit{Visible representative models}} \\ 
Gemini 2.5 Pro
& 69.63 & 43.00 & 21.43 & 16.67 & 29.92    & 29.43 & 39.80 &  48.15 \\
Qwen2.5-VL-32B
& 86.16 & 39.91 &  2.63 &  7.41 & 37.24    & 25.23 & 50.86  & 40.74 \\
TimeChat-Online-7B
& 98.51 & 33.33 &  0.00 &  0.00 & 29.72  & 26.01 & 37.84 &  31.63 \\
\midrule
\multicolumn{4}{l}{\textit{Blind representative models}} \\ 
Gemini 2.5 Pro blind
& 40.24 & 22.35 & 37.33 & 14.10 & 19.37 & 26.16 & 23.60 & 25.32 \\
Qwen2.5-VL-32B blind
& 00.00 & 29.50 & 0.00 & 0.00 & 27.80 & 26.99  & 31.58 & 26.30 \\
TimeChat-Online-7B blind
& 90.63 & 28.38 & 0.00 & 3.70 & 18.01 & 26.87 & 27.45 & 25.54 \\






\bottomrule
\end{tabular}
\end{table*}

%% file: supplement_materials/table_absurd_when_wrong.tex

\definecolor{future}{RGB}{218,227,243}      
\definecolor{futurelight}{RGB}{240,244,250} 
\definecolor{unpredlight}{RGB}{252,228,236} 
\providecommand{\hc}[2]{\cellcolor{#1}\hspace{1.5pt}#2\hspace{1.5pt}}

\providecommand{\MCQTaskBlock}[2]{}
\renewcommand{\MCQTaskBlock}[2]{%
  \cellcolor{#1}\makebox[0.148\textwidth][c]{#2}%
}
\providecommand{\MCQSubBlock}[2]{}
\renewcommand{\MCQSubBlock}[2]{%
  \cellcolor{#1}\makebox[0.074\textwidth][c]{#2}%
}
\providecommand{\MCQNumBlock}[1]{}
\renewcommand{\MCQNumBlock}[1]{%
  \makebox[0.074\textwidth][c]{#1}%
}

\noindent\begin{minipage}{\textwidth}
\centering
\tiny
\setlength{\tabcolsep}{0pt}
\renewcommand{\arraystretch}{1.05}
\refstepcounter{table}\label{tab:mcq_reshuffle}
\vspace{0.4em}
\textbf{Table~\thetable. MCQ option-order robustness and strong-distractor validation.} Conditional distribution of models selecting strong distractors under the original shuffled and re-shuffled MCQ options.

\begin{tabular}{@{}p{0.26\textwidth}p{0.074\textwidth}p{0.074\textwidth}p{0.074\textwidth}p{0.074\textwidth}p{0.074\textwidth}p{0.074\textwidth}p{0.074\textwidth}p{0.074\textwidth}p{0.074\textwidth}p{0.074\textwidth}@{}}
\toprule

\multirow{2}{*}{\makecell[l]{Model}} &
\multicolumn{2}{@{}c@{}}{\MCQTaskBlock{future}{\makecell[c]{Now narr.}}} &
\multicolumn{2}{@{}c@{}}{\MCQTaskBlock{future}{\makecell[c]{Sh.anticip.}}} &
\multicolumn{2}{@{}c@{}}{\MCQTaskBlock{future}{\makecell[c]{Ms.anticip.}}} &
\multicolumn{2}{@{}c@{}}{\MCQTaskBlock{future}{\makecell[c]{Sh.rtrv.}}} &
\multicolumn{2}{@{}c@{}}{\MCQTaskBlock{future}{\makecell[c]{Ms.rtrv.}}} \\
\cmidrule(lr){2-3}
\cmidrule(lr){4-5}
\cmidrule(lr){6-7}
\cmidrule(lr){8-9}
\cmidrule(lr){10-11}

& \MCQSubBlock{futurelight}{orig.}
& \MCQSubBlock{unpredlight}{\makecell[c]{re-shuf.}}
& \MCQSubBlock{futurelight}{orig.}
& \MCQSubBlock{unpredlight}{\makecell[c]{re-shuf.}}
& \MCQSubBlock{futurelight}{orig.}
& \MCQSubBlock{unpredlight}{\makecell[c]{re-shuf.}}
& \MCQSubBlock{futurelight}{orig.}
& \MCQSubBlock{unpredlight}{\makecell[c]{re-shuf.}}
& \MCQSubBlock{futurelight}{orig.}
& \MCQSubBlock{unpredlight}{\makecell[c]{re-shuf.}} \\
\midrule

Gemini 2.5 Pro
& \MCQNumBlock{98.20} & \MCQNumBlock{97.02}
& \MCQNumBlock{98.34} & \MCQNumBlock{98.01}
& \MCQNumBlock{96.79} & \MCQNumBlock{97.35}
& \MCQNumBlock{99.42} & \MCQNumBlock{99.77}
& \MCQNumBlock{99.05} & \MCQNumBlock{98.05} \\

Claude Sonnet 4
& \MCQNumBlock{91.58} & \MCQNumBlock{90.49}
& \MCQNumBlock{93.97} & \MCQNumBlock{91.42}
& \MCQNumBlock{94.68} & \MCQNumBlock{90.09}
& \MCQNumBlock{97.80} & \MCQNumBlock{96.84}
& \MCQNumBlock{90.68} & \MCQNumBlock{88.07} \\

\midrule
Qwen2.5-VL-72B
& \MCQNumBlock{98.47} & \MCQNumBlock{97.72}
& \MCQNumBlock{99.07} & \MCQNumBlock{99.06}
& \MCQNumBlock{99.50} & \MCQNumBlock{99.67}
& \MCQNumBlock{99.78} & \MCQNumBlock{100.00}
& \MCQNumBlock{98.91} & \MCQNumBlock{99.09} \\

Qwen2.5-VL-32B
& \MCQNumBlock{99.00} & \MCQNumBlock{99.28}
& \MCQNumBlock{99.20} & \MCQNumBlock{98.74}
& \MCQNumBlock{98.39} & \MCQNumBlock{98.78}
& \MCQNumBlock{100.00} & \MCQNumBlock{100.00}
& \MCQNumBlock{99.38} & \MCQNumBlock{99.79} \\

Qwen2.5-VL-7B
& \MCQNumBlock{98.03} & \MCQNumBlock{97.39}
& \MCQNumBlock{97.81} & \MCQNumBlock{96.74}
& \MCQNumBlock{99.31} & \MCQNumBlock{99.16}
& \MCQNumBlock{96.98} & \MCQNumBlock{95.86}
& \MCQNumBlock{98.49} & \MCQNumBlock{97.47} \\

Video-LLaVA 7B
& \MCQNumBlock{68.55} & \MCQNumBlock{66.18}
& \MCQNumBlock{69.86} & \MCQNumBlock{69.97}
& \MCQNumBlock{62.60} & \MCQNumBlock{64.74}
& \MCQNumBlock{67.15} & \MCQNumBlock{70.58}
& \MCQNumBlock{66.99} & \MCQNumBlock{65.52} \\

\midrule
TimeChat-Online-7B
& \MCQNumBlock{96.30} & \MCQNumBlock{97.15}
& \MCQNumBlock{96.84} & \MCQNumBlock{96.53}
& \MCQNumBlock{96.74} & \MCQNumBlock{97.17}
& \MCQNumBlock{99.34} & \MCQNumBlock{99.33}
& \MCQNumBlock{89.10} & \MCQNumBlock{90.35} \\

ROI-TimeChat-Online
& \MCQNumBlock{94.80} & \MCQNumBlock{94.92}
& \MCQNumBlock{84.50} & \MCQNumBlock{84.63}
& \MCQNumBlock{96.87} & \MCQNumBlock{87.18}
& \MCQNumBlock{98.00} & \MCQNumBlock{98.59}
& \MCQNumBlock{90.62} & \MCQNumBlock{89.44} \\

Flash-VStream
& \MCQNumBlock{93.52} & \MCQNumBlock{96.45}
& \MCQNumBlock{94.07} & \MCQNumBlock{96.79}
& \MCQNumBlock{94.53} & \MCQNumBlock{97.67}
& \MCQNumBlock{97.34} & \MCQNumBlock{97.84}
& \MCQNumBlock{97.31} & \MCQNumBlock{98.98} \\

LLaVA-OV1.5
& \MCQNumBlock{98.74} & \MCQNumBlock{98.48}
& \MCQNumBlock{95.57} & \MCQNumBlock{94.54}
& \MCQNumBlock{97.45} & \MCQNumBlock{99.16}
& \MCQNumBlock{98.80} & \MCQNumBlock{98.35}
& \MCQNumBlock{96.65} & \MCQNumBlock{96.98} \\

\midrule
\mbox{TimeChat-Online-7B(SFT)}
& \MCQNumBlock{100.00} & \MCQNumBlock{99.46}
& \MCQNumBlock{98.97} & \MCQNumBlock{97.73}
& \MCQNumBlock{99.78} & \MCQNumBlock{99.34}
& \MCQNumBlock{100.00} & \MCQNumBlock{100.00}
& \MCQNumBlock{96.38} & \MCQNumBlock{98.42} \\

\mbox{ROI-TimeChat-Online(SFT)}
& \MCQNumBlock{100.00} & \MCQNumBlock{100.00}
& \MCQNumBlock{90.96} & \MCQNumBlock{91.40}
& \MCQNumBlock{99.79} & \MCQNumBlock{99.02}
& \MCQNumBlock{99.82} & \MCQNumBlock{99.77}
& \MCQNumBlock{98.05} & \MCQNumBlock{98.42} \\

\bottomrule
\end{tabular}
\end{minipage}

%% file: supplement_materials/table_open.tex
\begin{table*}[t]
\centering
\scriptsize
\setlength{\tabcolsep}{1.5pt}  
\renewcommand{\arraystretch}{1.15}
\caption{\textbf{Detailed results of Now narration}: open-ended accuracy on verbs, nouns, and actions, and MCQ accuracy.}
\label{tab:now_narration}
\resizebox{1.0\textwidth}{!}{%
\begin{tabular}{lcccc}
\toprule

\noalign{\global\setlength{\tabcolsep}{0pt}}
\multirow{3}{*}{\makecell[l]{model}} &
\multicolumn{4}{c}{\hc{now}{Now narration}} \\
\cmidrule(lr){2-5}

& \multicolumn{3}{c}{\hc{nowlight}{Open-ended mode}} &
\multicolumn{1}{c}{\hc{mcqlight}{MCQ mode}} \\
\cmidrule(lr){2-5}

& \hc{nowlight}{Verb acc}
& \hc{nowlight}{Noun acc}
& \hc{nowlight}{Action acc}
& \hc{mcqlight}{MCQ acc} \\
\midrule

\noalign{\global\setlength{\tabcolsep}{1.5pt}}

Gemini 2.5 Pro & 10.57 & 21.43 & 4.16 & 43.00 \\

\makecell[l]{Qwen 2.5-VL-\\32B} & 9.65 & 13.82 & 2.25 & 39.91 \\

TimeChat-Online-7B & 1.17 & 3.82 & 0.34 & 33.33 \\

\bottomrule
\end{tabular}
}%
\end{table*}

%% file: supplement_materials/table_memory_proxy.tex

\definecolor{past}{RGB}{226,240,217}     
\definecolor{current}{RGB}{255,242,204}  
\definecolor{future}{RGB}{218,227,243}   

\definecolor{pastlight}{RGB}{241,247,237}
\definecolor{currentlight}{RGB}{255,249,229}
\definecolor{futurelight}{RGB}{240,244,250}

\providecommand{\hc}[2]{\cellcolor{#1}\hspace{1.5pt}#2\hspace{1.5pt}}

\newcommand{\ModelBlock}[2]{%
  {\setlength{\fboxsep}{0pt}%
  \colorbox{#1}{%
    \parbox[c][4.9em][c]{0.135\textwidth}{\raggedright\hspace{1pt}#2}%
  }}%
}
\newcommand{\MemBlock}[2]{%
  \cellcolor{#1}\makebox[0.135\textwidth][l]{\hspace{1pt}#2}%
}
\newcommand{\HeadBlock}[2]{%
  \cellcolor{#1}\makebox[0.146\textwidth][c]{#2}%
}
\newcommand{\NumBlock}[1]{%
  \makebox[0.146\textwidth][c]{#1}%
}

\par\vspace{0.8\baselineskip}
\noindent\begin{minipage}{\textwidth}
\centering
\tiny
\noindent\begin{minipage}{\textwidth}
\centering
\tiny
\setlength{\tabcolsep}{0pt}
\renewcommand{\arraystretch}{1.0}
\refstepcounter{table}\label{tab:memory_proxy_results}
\vspace{-0.2em}
\textbf{Table~\thetable. Memory proxy sensitivity.} MCQ accuracy under different memory context horizons.
\begin{tabular}{@{}p{0.155\textwidth}p{0.115\textwidth}p{0.146\textwidth}p{0.146\textwidth}p{0.146\textwidth}p{0.146\textwidth}p{0.146\textwidth}@{}}
\toprule

\makecell[l]{Model} &
\makecell[l]{Memory\\config} &
\HeadBlock{future}{\makecell[c]{Now\\narration}} &
\HeadBlock{future}{\makecell[c]{Sh.\\anticipation}} &
\HeadBlock{future}{\makecell[c]{Ms.\\anticipation}} &
\HeadBlock{future}{\makecell[c]{Sh.\\retrieval}} &
\HeadBlock{future}{\makecell[c]{Ms.\\retrieval}} \\
\midrule

\multirow{4}{*}{\ModelBlock{current}{Gemini2.5\\Pro}}
& \MemBlock{currentlight}{D10S10} & \NumBlock{37.91} & \NumBlock{31.43} & \NumBlock{31.24} & \NumBlock{41.81} & \NumBlock{40.42} \\
 \cmidrule(lr){2-7}
& \MemBlock{currentlight}{D10S20} & \NumBlock{40.22} & \NumBlock{33.61} & \NumBlock{29.84} & \NumBlock{41.94} & \NumBlock{43.60} \\
\cmidrule(lr){2-7}
& \MemBlock{currentlight}{D10S30} & \NumBlock{40.60} & \NumBlock{34.66} & \NumBlock{30.54} & \NumBlock{42.87} & \NumBlock{40.23} \\
\cmidrule(lr){2-7}
& \MemBlock{currentlight}{D10S40} & \NumBlock{42.04} & \NumBlock{31.27} & \NumBlock{28.52} & \NumBlock{42.42} & \NumBlock{40.00} \\
\midrule

\multirow{4}{*}{\ModelBlock{past}{Qwen2.5-\\VL-7B}}
 & \MemBlock{pastlight}{D10S10} & \NumBlock{31.78} & \NumBlock{27.62} & \NumBlock{27.74} & \NumBlock{38.65} & \NumBlock{26.10} \\
 \cmidrule(lr){2-7}
& \MemBlock{pastlight}{D10S20} & \NumBlock{31.63} & \NumBlock{29.26} & \NumBlock{27.86} & \NumBlock{37.74} & \NumBlock{28.27} \\
\cmidrule(lr){2-7}
& \MemBlock{pastlight}{D10S30} & \NumBlock{32.07} & \NumBlock{28.24} & \NumBlock{29.35} & \NumBlock{37.44} & \NumBlock{27.60} \\
\cmidrule(lr){2-7}
& \MemBlock{pastlight}{D10S40} & \NumBlock{32.18} & \NumBlock{29.14} & \NumBlock{28.73} & \NumBlock{38.78} & \NumBlock{28.52} \\

\bottomrule
\end{tabular}
\end{minipage}
\end{minipage}
\par\vspace{0.8\baselineskip}

%% file: supplement_materials/table_cross_scenario.tex

\definecolor{future}{RGB}{218,227,243}      
\definecolor{futurelight}{RGB}{240,244,250} 
\definecolor{unpredlight}{RGB}{252,228,236} 
\providecommand{\hc}[2]{\cellcolor{#1}\hspace{1.5pt}#2\hspace{1.5pt}}

\providecommand{\SFTTaskBlock}[2]{}
\renewcommand{\SFTTaskBlock}[2]{%
  \cellcolor{#1}\makebox[0.148\textwidth][c]{#2}%
}
\providecommand{\SFTSubBlock}[2]{}
\renewcommand{\SFTSubBlock}[2]{%
  \cellcolor{#1}\makebox[0.074\textwidth][c]{#2}%
}
\providecommand{\SFTNumBlock}[1]{}
\renewcommand{\SFTNumBlock}[1]{%
  \makebox[0.074\textwidth][c]{#1}%
}

\smallskip
\noindent\begin{minipage}{\textwidth}
\centering
\tiny
\setlength{\tabcolsep}{0pt}
\renewcommand{\arraystretch}{1.12}
\refstepcounter{table}\label{tab:sft_transfer}
\vspace{-0.1em}
\textbf{Table~\thetable. SFT transfer analysis.} Performance of TimeChat SFT variants under source-to-target task settings.
\begin{tabular}{@{}p{0.26\textwidth}p{0.074\textwidth}p{0.074\textwidth}p{0.074\textwidth}p{0.074\textwidth}p{0.074\textwidth}p{0.074\textwidth}p{0.074\textwidth}p{0.074\textwidth}p{0.074\textwidth}p{0.074\textwidth}@{}}
\toprule

\multirow{2}{*}{\makecell[l]{Model$\backslash$task}} &
\multicolumn{2}{@{}c@{}}{\SFTTaskBlock{future}{\makecell[c]{Now narr.}}} &
\multicolumn{2}{@{}c@{}}{\SFTTaskBlock{future}{\makecell[c]{Sh.anticip.}}} &
\multicolumn{2}{@{}c@{}}{\SFTTaskBlock{future}{\makecell[c]{Ms.anticip.}}} &
\multicolumn{2}{@{}c@{}}{\SFTTaskBlock{future}{\makecell[c]{Sh.rtrv.}}} &
\multicolumn{2}{@{}c@{}}{\SFTTaskBlock{future}{\makecell[c]{Ms.rtrv.}}} \\
\cmidrule(lr){2-3}
\cmidrule(lr){4-5}
\cmidrule(lr){6-7}
\cmidrule(lr){8-9}
\cmidrule(lr){10-11}

& \SFTSubBlock{futurelight}{A$\rightarrow$B}
& \SFTSubBlock{unpredlight}{B$\rightarrow$B}
& \SFTSubBlock{futurelight}{A$\rightarrow$B}
& \SFTSubBlock{unpredlight}{B$\rightarrow$B}
& \SFTSubBlock{futurelight}{A$\rightarrow$B}
& \SFTSubBlock{unpredlight}{B$\rightarrow$B}
& \SFTSubBlock{futurelight}{A$\rightarrow$B}
& \SFTSubBlock{unpredlight}{B$\rightarrow$B}
& \SFTSubBlock{futurelight}{A$\rightarrow$B}
& \SFTSubBlock{unpredlight}{B$\rightarrow$B} \\
\midrule

TimeChat-Online-7B (SFT)
& \SFTNumBlock{36.60} & \SFTNumBlock{37.43}
& \SFTNumBlock{50.14} & \SFTNumBlock{50.89}
& \SFTNumBlock{40.81} & \SFTNumBlock{38.11}
& \SFTNumBlock{44.87} & \SFTNumBlock{45.18}
& \SFTNumBlock{42.24} & \SFTNumBlock{43.23} \\

ROI-TimeChat-Online (SFT)
& \SFTNumBlock{40.35} & \SFTNumBlock{41.36}
& \SFTNumBlock{51.59} & \SFTNumBlock{51.44}
& \SFTNumBlock{38.22} & \SFTNumBlock{39.07}
& \SFTNumBlock{43.94} & \SFTNumBlock{44.39}
& \SFTNumBlock{38.42} & \SFTNumBlock{38.14} \\

\bottomrule
\end{tabular}
\end{minipage}

%% file: supplement_materials/related_work.tex
\section {Comparison with Related Benchmarks}

\newcommand{\cmark}{\ding{51}}%
\newcommand{\xmark}{\ding{55}}%
\definecolor{mygray}{gray}{0.9}

\definecolor{mygray}{gray}{0.9}

\begin{table}[t]
  \centering
  \caption{Comparison of \textbf{EgoSAT} with representative ego/exo-centric VQA benchmarks under both offline and online/streaming settings. We summarize key dataset properties including the number of samples, egocentric perspective, task form (MCQ or open-ended), annotation pipeline, and temporal reasoning hierarchy. Temporal labels denote whether questions require \textit{retrospective (retro.)}, \textit{present (pres.)}, or \textit{prospective (pros.)} reasoning. Only \textbf{EgoSAT} explicitly evaluates answerability, a necessary capability for reasoning under temporal uncertainty in streaming settings. For clarity, note that human review for quality assurance is adopted by all existing QA-generation pipelines, including ours.}
  \label{tab:benchmark_comparison}

  \resizebox{\columnwidth}{!}{
    \begin{tabular}{@{}c l c c c c c c@{}}
    \toprule
    \textbf{Protocol} & 
    \textbf{Benchmark} & 
    \textbf{\# of samples} & 
    \textbf{Ego-centric} & 
    \makecell{\textbf{Task}\\\textbf{Form}} & 
    \textbf{Annotation} & 
    \makecell{\textbf{Temporal}\\\textbf{Hierarchy}} & 
    \makecell{\textbf{Answerability}\\\textbf{Eval.}} \\
    \midrule
    
      \multirow{16}{*}{Offline}
      & LongVideoBench~\cite{wu2024longvideobench} & 6,678 QA & \xmark & MCQ & Human & retro. + pres. + pros. & \xmark \\
      & LVBench~\cite{wang2025lvbench} & 1,549 QA & \xmark & MCQ & Human & retro. + pres. + pros. & \xmark \\
      & MLVU~\cite{zhou2025mlvu} & 3,102 QA & \xmark & Both & LLM+Human & retro. + pres. + pros. & \xmark \\
      & EgoVQA~\cite{fan2019egovqa} & 600 QA & \cmark & Both & Human & pres. & \xmark \\
      & EgoMemoria~\cite{ye2025mmego} & 7,026 QA & \cmark & MCQ & LLM & retro. & \xmark \\
      & QAEgo4D~\cite{Barmann2022ego4dqa} & 14,513 QA & \cmark & Open & Human & retro. & \xmark \\
      & AssistQ~\cite{benita2022assistq} & 531 QA & \cmark & MCQ & Human & pres. + pros. & \xmark \\
      & EgoThink~\cite{Cheng2023EgoThinkEF} & 700 QA & \cmark & Open & Human & pres. + pros. & \xmark \\
      & EgoTaskQA~\cite{jia2022egotaskqa} & 40,000 QA & \cmark & Open & Human & retro. + pres. + pros. & \xmark \\
      & EOC-Bench~\cite{yuan2025eoc} & 3,277 QA & \cmark & Both & Human & retro. + pres. + pros. & \xmark \\
      & VideoMindPalace~\cite{huang2025building} & 1,800 QA & \cmark & Both & LLM+Human & retro. + pres. & \xmark \\
      & EgoGazeVQA~\cite{Peng2025InTE} & 1,757 QA & \cmark & MCQ & MLLM+Human & retro. + pres. & \xmark \\
      & EgoSchema~\cite{Mangalam2023EgoSchemaAD} & 5,063 QA & \cmark & MCQ & LLM+Human & retro. + pres. & \xmark \\
      & EgoPlan~\cite{Chen2023EgoPlanBenchBM} & 4,939 QA & \cmark & MCQ & LLM+Human & pros. & \xmark \\
      & EgoLifeQA~\cite{yang2025egolife} & 3,000 QA & \cmark & MCQ & LLM+Human & retro. + pres. + pros. & \xmark \\

      \midrule

      \multirow{5}{*}{Streaming}
      & EgoTextVQA~\cite{Zhou2025EgoTextVQATE} & 7,064 QA & \cmark & Open & MLLM+Human & retro. + pres. + pros. & \xmark \\
      & OVO-Bench~\cite{niu2025ovo} & 2,814 QA & \xmark & Both & MLLM+Human & retro. + pres. + pros. & \xmark \\
      & ESTP~\cite{zhangeyes} & 2,264 QA & \cmark & Open & LLM+Human & pros. & \xmark \\
      & StreamingBench~\cite{lin2024streamingbench} & 4,500 QA & \xmark & MCQ & LLM+Human & retro. + pres. + pros. & \xmark \\
      & SVBench~\cite{yangsvbench} & 49,979 QA & \xmark & Open & LLM+Human & retro. + pres. + pros. & \xmark \\
      & OmniMMI~\cite{wang2025omnimmi} & 2,290 QA & \cmark & Open & Human & retro. + pres. + pros. & \xmark \\
      & \cellcolor{mygray}Our Benchmark & \cellcolor{mygray}4,800 QA & \cellcolor{mygray}\cmark & \cellcolor{mygray}Both & \cellcolor{mygray}Human & \cellcolor{mygray}retro. + pres. + pros. & \cellcolor{mygray}\cmark \\
      \bottomrule
    \end{tabular}
  }
\end{table}

Tab.~\ref{tab:benchmark_comparison} situates our \textbf{EgoSAT} within the landscape of existing ego/exo-centric VQA benchmarks. Notably, most existing benchmarks focus on the offline setting and do not jointly evaluate retrospective, present, and prospective reasoning. Among the limited streaming benchmarks, the most relevant prior work is EWO, which focuses on open-ended question answering but does not explicitly model past or future temporal reasoning. OmniMMI is another recent benchmark for multi-modal interaction in streaming video contexts, covering streaming video understanding and proactive reasoning with egocentric videos; however, it targets general OmniLLM-style audio-visual interaction and does not explicitly evaluate answerability or confidence under temporal uncertainty. In contrast, EgoSAT is the only benchmark that explicitly evaluates answerability by incorporating both confidence and predictability into the benchmark design.

\section {Limitations and Future Work}
\noindent\textbf{Limitations}.\ Though we provide the first benchmark to systematically study egocentric streaming interaction understanding across temporal hierarchies, most of our evaluations focus on adapting existing offline models to the streaming setting. Currently, relatively few works have investigated native online streaming models, and the limited existing approaches often exhibit weak temporal reasoning capabilities due to aggressive memory caching strategies and constrained model sizes. In our experiments, TimeChat-Online serves as the most feasible baseline for streaming evaluation. In addition, although we conduct supervised fine tuning to improve performance on multiple choice and state switching tasks, enabling models to properly reason about answerability, particularly through calibrated confidence estimation and reasoning about event predictability, remains an open challenge.

\noindent\textbf{Future Work}.\ In future work, we plan to explore improved caching mechanisms for online streaming models. In particular, we will investigate dynamic caching strategies that incorporate region of interest selection while explicitly modeling temporal dependencies, enabling models to retain temporally informative observations under limited memory budgets. In addition, we will study reinforcement learning based training to guide models in reasoning about answerability, with a focus on learning calibrated confidence and recognizing inherently unpredictable future events.